\documentclass[10pt,twocolumn,letterpaper]{article}

\PassOptionsToPackage{table}{xcolor}
\usepackage[pagenumbers]{cvpr} 

\definecolor{cvprblue}{rgb}{0.21,0.49,0.74}
\usepackage[pagebackref,breaklinks,colorlinks,allcolors=cvprblue]{hyperref}
\usepackage{multirow}
\usepackage{amssymb}
\usepackage{pifont}
\usepackage{array}
\usepackage{booktabs} 
\usepackage{float}
\usepackage{graphicx}

\def\paperID{6100} 
\def\confName{CVPR}
\def\confYear{2025}

\title{LongVALE: Vision-Audio-Language-Event Benchmark Towards \\ Time-Aware Omni-Modal Perception of Long Videos}

\author{Tiantian Geng$^{1,2}$, Jinrui Zhang$^{1}$, Qingni Wang$^{3}$, Teng Wang$^{1,4}$, Jinming Duan$^{2,5*}$, Feng Zheng$^{1*}$\\
$^1$Southern University of Science and Technology \ 
$^2$University of Birmingham\\
$^3$University of Electronic Science and Technology of China \\
$^4$The University of Hong Kong \
$^5$University of Manchester \\
{\tt\small gengtiantian97@gmail.com\ zhangjr2018@mail.sustech.edu.cn }
}

\begin{document}
\twocolumn[{%
\renewcommand\twocolumn[1][]{#1}%
\maketitle

\begin{center}
    \centering
  \vspace{-6mm}
    \captionsetup{type=figure}
      \includegraphics[width=0.97\linewidth]{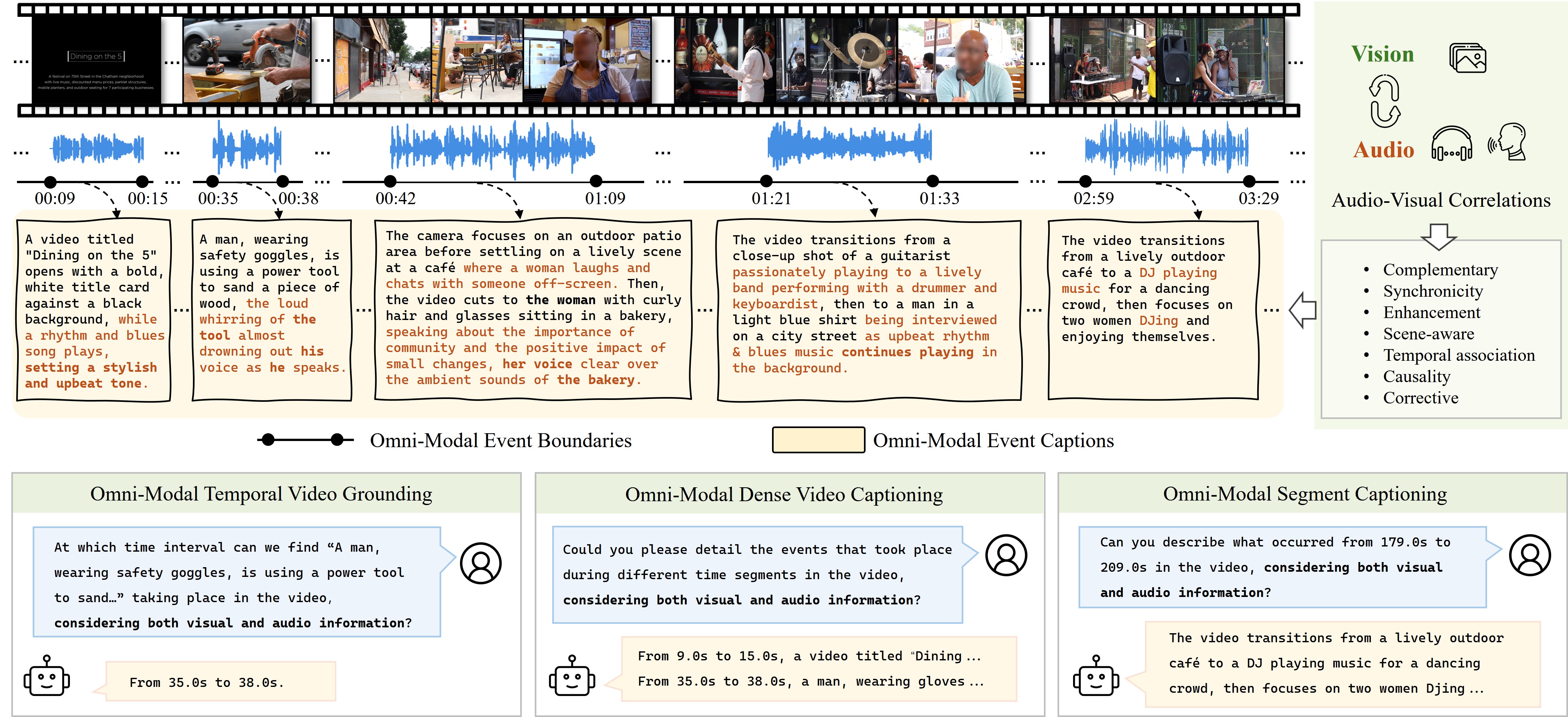}
    \captionof{figure}{We introduce LongVALE, the first-ever omni-modality long video benchmark,  
    offering precise temporal boundaries and captions for omni-modal events integrating visual, audio, and speech information. The captions feature audio-visual correlations to enhance cross-modal learning. Besides, we extend three fine-grained video tasks to the omni-modality domain, enabling omni-perception of long videos.}
  \label{fig:fig1}
\end{center}%
}]
\let\thefootnote\relax\footnotetext{{$*$ Corresponding co-authors}}
\begin{abstract}
Despite impressive advancements in video understanding, most efforts remain limited to coarse-grained or visual-only video tasks. However, real-world videos encompass omni-modal information (vision, audio, and speech) with a series of events forming a cohesive storyline. 
The lack of multi-modal video data with fine-grained event annotations and the high cost of manual labeling are major obstacles to comprehensive omni-modality video perception.
To address this gap, we propose an automatic pipeline consisting of high-quality multi-modal video filtering, semantically coherent omni-modal event boundary detection, and cross-modal correlation-aware event captioning. 
In this way, we present LongVALE, the first-ever Vision-Audio-Language Event understanding benchmark comprising 105K omni-modal events with precise temporal boundaries and detailed relation-aware captions within 8.4K high-quality long videos.
Further, we build a baseline that leverages LongVALE to enable video large language models (LLMs) for omni-modality fine-grained temporal video understanding for the first time.
Extensive experiments demonstrate the effectiveness and great potential of LongVALE in advancing comprehensive multi-modal video understanding. The dataset and code are available at \url{https://ttgeng233.github.io/LongVALE/}.
\end{abstract}
\vspace{-5.5mm}    
\section{Introduction}
\label{sec:intro}
\begin{table*}
  \centering
  \resizebox{\linewidth}{!}{ 
  \begin{tabular}{l|cccc|ccc|cc|c}
    \toprule
    Dataset & Annotation & \#Videos & Avg. video len & \#Avg. event & Vision & Audio & Subtitle & Captions & Timestamps & A-V Correlations  \\
    \midrule
    
    InternVid~\cite{wang2023internvid} & G & 234M & 11.7s & 1 & \checkmark & \ding{55}& \ding{55}& V & - & \ding{55}   \\
    Panda-70M~\cite{chen2024panda} & G & 70.8M & 8.5s & 1 & \checkmark & \ding{55}& \ding{55}& V & - & \ding{55} \\
    AudioCaps~\cite{kim2019audiocaps} & M & 51.3K & 10s & 1 & \ding{55} & \checkmark & \ding{55} & A & - & \ding{55} \\
    WavCaps~\cite{mei2024wavcaps} & G & 403K & 67.6s & 1 & \ding{55} & \checkmark & \ding{55} & A & - & \ding{55}\\
    ACAV~\cite{lee2021acav100m} & G & 100M & 10s & 1 & \checkmark & \checkmark & \ding{55} & - & - & \ding{55} \\
    VALOR~\cite{liu2024valor} & M & 1.18M & 10s & 1 & \checkmark & \checkmark & \ding{55} & VA & - & \ding{55}\\
    VAST~\cite{chen2023vast} & G & 27M & 5$\sim$30s & 1 & \checkmark & \checkmark & \checkmark & VAS & - & \ding{55} \\
    \midrule
    AVEL~\cite{tian2018audio} & M & 4,143 & 10s & 1 & \checkmark & \checkmark & \ding{55} & - & VA & \ding{55}\\
    UnAV-100~\cite{geng2023dense} & M & 10,790 & 42.1s & 2.8 & \checkmark & \checkmark & \ding{55} & - & VA & \ding{55} \\
    ActivityNet Caps~\cite{krishna2017dense}& M & 20K & 180s & 3.7 &  \checkmark & \ding{55} & \ding{55} & V & V & \ding{55}\\
    Charades-STA~\cite{gao2017tall} & M & 10K & 30s & 1.6 &  \checkmark & \ding{55} & \ding{55} & V & V & \ding{55}\\
    \midrule
    \textbf{LongVALE (Ours)} & G+M & 8,411 & 235s & 12.6 & \checkmark & \checkmark & \checkmark & VAS & VAS  & \checkmark\\
    \bottomrule
  \end{tabular}}
  \vspace{-2mm}
  \caption{Comparison of LongVALE with previous related benchmarks. G: generated. M: manual. V: visual. A: audio. S: speech. }
  \label{tab:dataset_compar}
  \vspace{-5mm}
\end{table*}

As the volume of videos on social media platforms grows exponentially, video understanding~\cite{krishna2017dense,gao2017tall,wang2023internvid,zhang2023video} has emerged as a vital research area in artificial intelligence.
When watching videos, such as daily vlogs or tutorials lasting several minutes, viewers need to integrate visual and auditory information and associate multiple events to fully comprehend the content. 
An ideal intelligent video agent should imitate it, capable of both cross-modal reasoning and fine-grained temporal understanding. 
However, current research is limited to coarse-grained tasks (\eg, video retrieval/captioning~\cite{wang2023internvid,liu2024valor}) or visual-only fine-grained tasks (\eg, temporal grounding/dense captioning~\cite{krishna2017dense, gao2017tall}), remaining far from enough to achieve both the capabilities.

A significant barrier to this advancement is the absence of a high-quality video dataset with omni-modality (vision, audio, and speech) and fine-grained temporal annotations. 
As seen in Tab.~\ref{tab:dataset_compar}, current benchmarks either contain only global captions for short video/audio clips~\cite{wang2023internvid, chen2024panda, kim2019audiocaps,mei2024wavcaps,liu2024valor,chen2023vast}, offer visual-only multi-event annotations~\cite{krishna2017dense,gao2017tall}, or possess multi-modal events but lack detailed captions~\cite{tian2018audio,geng2023dense}.
Moreover, creating such a dataset
poses significant challenges, as identifying temporal boundaries and producing detailed event captions by integrating information from various modalities are difficult and time-consuming, even for human annotators.  

In this work, we propose an efficient and scalable annotation pipeline, capable of generating temporal boundaries and detailed captions for omni-modal events (\ie, integrating vision, audio, and speech) within arbitrary multi-modal long videos.
Our pipeline includes three distinct aspects: 
1) \textit{High-quality video filtering for rich audio-visual semantics and temporal dynamics.}
The filtered videos showcase rich dynamic visual scenes paired with diverse audio types, \eg, instruments playing, people laughing, and tools whirring as in Fig.~\ref{fig:fig1}, contrasting with the dense narration in prior datasets~\cite{chen2024panda, chen2023vast}. 
2) \textit{Omni-modal event boundary detection for semantic coherence in both visual and audio scenes.} Unlike previous works~\cite{krishna2017dense,gao2017tall} that only identify visual event boundaries, we determine omni-modal event boundaries utilizing both visual and audio cues. This prevents audio scenes from being cut off, avoiding the loss of critical information.
3) \textit{Omni-modal event captioning emphasizing audio-visual correlation reasoning.} Instead of simple concatenation~\cite{chen2023vast,liu2024valor}, we fully integrate modality-specific information (vision, audio, speech) and explicitly reason about their correlations to enhance cross-modal understanding.
For example, in Fig.~\ref{fig:fig1}, the visible man using the tool with loud whirring reflects audio-visual \textit{synchronicity}, and the woman's speech crucially \textit{complements} the visual scene. 

Based on the pipeline, we construct LongVALE, the first-ever benchmark for omni-modality fine-grained video understanding. It comprises 8.4K long videos containing 105K omni-modal events annotated with high-quality temporal boundaries and correlation-aware captions. 
Notably, it features a longer average video length (235 seconds) and more events (12.6 per video) compared to existing benchmarks, along with its unique omni-modal event boundaries and captions with audio-visual correlations (seen in Tab.~\ref{tab:dataset_compar}).

Building upon LongVALE, we present LongVALE-LLM, a multi-modal video LLM, capable of both cross-modal reasoning and fine-grained temporal understanding. Further, we extend three fine-grained video tasks (\ie, temporal video grounding, dense video captioning, and segment captioning) from a vision-oriented to a novel omni-modal setting.
During training, our dataset can serve as a highly valuable data source for both event boundary perception tuning~\cite{huang2024vtimellm} and instruction tuning.
Our experiments show that the model trained on our LongVALE dataset significantly outperforms existing video LLMs across all three tasks.
Moreover, we find that our model can surprisingly achieve superior performance on general audio-visual question answering (AVQA) tasks~\cite{alamri2019audio,li2022learning} in a zero-shot manner, even with significantly less data compared to other LLM-based methods trained on million-scale data. 
It highlights the effectiveness and great promise of our dataset in diving forward comprehensive multi-modal video understanding.
\begin{figure*}[t]
  \centering
  \setlength{\abovecaptionskip}{1.0mm}
   \includegraphics[width=1.0\linewidth]{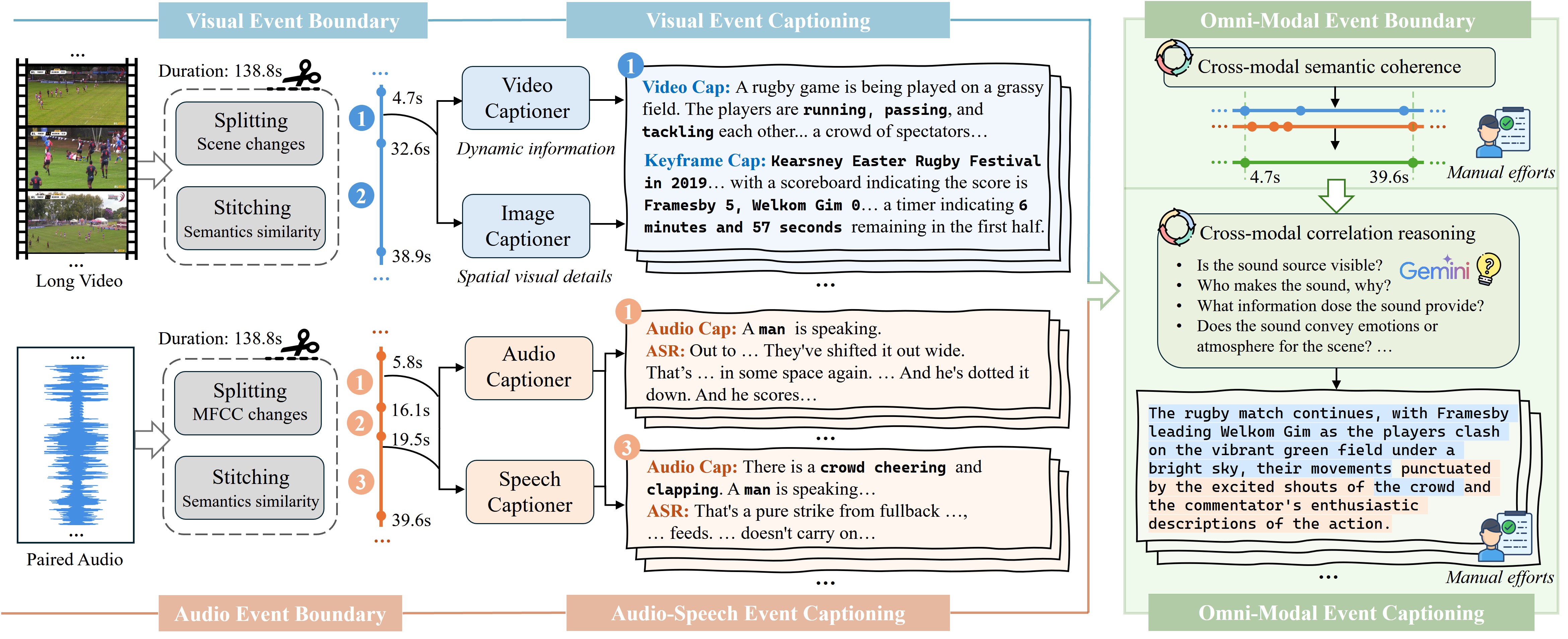}
   \caption{The pipeline for high-quality omni-modality fine-grained data generation. 
   It starts by detecting visual and audio event boundaries based on their distinct properties. 
   Next, we generate detailed captions for each video and audio event enhanced by keyframe and speech captions. 
   We then determine omni-modal event boundaries by maintaining the semantic integrity of single-modal events. Finally, omni-modal event captions are generated by audio-visual correlation reasoning, followed by manual refinement to ensure data's high quality.}
   \label{fig:dataset_pipeline}
     \vspace{-4mm}
\end{figure*}
Our contributions can be summarized as follows: 
\begin{itemize}
    \item We propose a novel scalable pipeline enabling the automatic generation of high-quality omni-modality fine-grained annotations for multi-modal long videos, significantly reducing manual annotation costs. 
    \item We introduce LongVALE, the first-ever benchmark providing omni-modal event temporal boundaries and cross-modal correlation-aware captions for 105K omni-modal events within 8.4K high-quality multi-modal long videos.
    \item We demonstrate that our LongVALE-trained model excels in both cross-modal reasoning and fine-grained temporal understanding, significantly outperforming existing video LLMs across all three omni-modal tasks and even achieving superior zero-shot results on general AVQA.
    
\end{itemize}

\section{Related Work}
\label{sec:related_work}
\vspace{-2mm}
\noindent \textbf{Multi-modal video benchmarks.}
Current research mainly focuses on building large-scale video/audio-language benchmarks. For instance, InternVid~\cite{wang2023internvid} and WavCaps~\cite{mei2024wavcaps} are web-scarped video-text and audio-text datasets composed of short clips. Moving forward, VALOR~\cite{liu2024valor} provides audio-visual captions and VAST~\cite{chen2023vast} further includes subtitles, but they just simply concatenate captions from different modalities, ignoring the cross-modal correlation reasoning.
These benchmarks offer only coarse-grained captions for short clips, which are unsuitable for fine-grained long video understanding.
{\color{black}Besides, some large-scale long video datasets~\cite{miech2019howto100m,zellers2021merlot,xue2022advancing} only use rough subtitles as annotations, failing to directly align with video content.} 
Moreover, fine-grained video benchmarks like ActivityNet Caps~\cite{krishna2017dense} and Charades-STA~\cite{gao2017tall} focus only on visual modality, while other audio-visual benchmarks like AVEL~\cite{tian2018audio} and UnAV-100~\cite{geng2023dense} provide temporal boundaries but lack rich captions. 
These limitations restrict models' abilities in both cross-modal reasoning and fine-grained temporal understanding for real-world videos. 
A detailed comparison with our LongVALE is shown in Tab.~\ref{tab:dataset_compar}.

\noindent \textbf{Fine-grained video understanding.}
To precisely locate and comprehend specific events in videos is crucial for video analysis, especially for untrimmed long videos.
Various fine-grained video tasks have been proposed, such as temporal video grounding~\cite{gao2017tall,luo2023towards} to identify temporal boundaries for a given text query, and dense video captioning~\cite{krishna2017dense,wang2021end,yang2023vid2seq}, demanding both temporal localization and captioning for all visual events.
Prior studies~\cite{gao2017tall,luo2023towards,yang2023vid2seq} handle each task separately on specialized datasets and some~\cite{lei2021detecting,liu2022umt,lin2023univtg} attempt to bridge several tasks in a unified model. 
Furthermore, recent video large language models (video LLMs) have shown promise in visual-only fine-grained video understanding ~\cite{ren2024timechat,huang2024vtimellm}.
In contrast, we aim to pioneer omni-modality fine-grained video understanding for a more holistic video comprehension.

\section{The LongVALE Benchmark}
To build LongVALE, we propose an efficient and scalable pipeline that includes high-quality multi-modal long video filtering (Sec.~\ref{sec:filter}), omni-modal event boundary detection (Sec.~\ref{sec:boundary}), and omni-modal event captioning with audio-visual correlation reasoning (Sec.~\ref{sec:3.3}). 
The annotation process is illustrated in Fig.~\ref{fig:dataset_pipeline}.
{\color{black}More details are in Appendix.}

\subsection{Data Collection and Filtering}
\label{sec:filter}
We source videos from ACAV-100M~\cite{lee2021acav100m}, which contains video clips with high audio-visual correspondence, covering a wide variety of topics. 
We download raw videos on YouTube without cutting to maintain the integrity of the video content, where the videos span 30 seconds to 10 minutes. 
Then, we design a filtering strategy to obtain high-quality videos containing rich visual and audio semantics, as well as temporal dynamic information. 

Firstly, we use metadata to filter out low-quality videos with resolutions below 360p and retain only those with English transcripts.
To collect videos with diverse sounds (\eg, \textit{people clapping/laughing}, \textit{dog barking}), we exclude those where speech dominates, defined as when subtitles cover over 95\% of the duration.
Further, we remove videos with static content (\ie, slide shows) using PySceneDetect~\cite{Pyscenedetect} to detect scenes.
If a scene's average frame difference is below a threshold, it is considered static, and videos with over 80\% static scenes are filtered to ensure diverse motion content.
Finally, we select videos with segments that have consistent audio-visual semantics. 
We split each video into 5-second visual and audio segments, then use C-MCR~\cite{wang2023connecting}, an audio-visual contrastive learning model, to compute similarity scores between segment embeddings. Videos with at least one segment having audio-visual similarity above 0.25 are retained.
This effectively filters out videos with irrelevant audio-visual signals, such as edited or overdubbed audio (\eg, background music and narration).

As a result, from a total of 100K raw videos collected from ACAV-100M, we obtained 8.4K videos, highlighting our high standard for video quality.

\subsection{Omni-Modal Event Boundary Detection}
\label{sec:boundary}
Existing video benchmarks~\cite{wang2023internvid,chen2024panda,krishna2017dense,gao2017tall} segment videos into events based solely on visual signals. 
However, audio is equally crucial for a complete video content understanding, and audio events often have boundaries that do not align with visual ones. 
For example, a scene shift from a stage performance to the audience or a news broadcast cutting from a live scene to the studio may disrupt the visual event, but the audio continues.
Thus, relying only on visual cues for event boundary detection can severely break the semantic coherence of the audio.
To solve the issue, we propose to detect omni-modal event boundaries for the first time, considering both visual and audio scene boundaries. 

\noindent \textbf{Visual event boundary.}
Using only visual cues, we apply a two-stage detection method~\cite{chen2024panda} which includes splitting basic visual scenes and then merging semantically similar ones.
Notably, we refine the previous method to handle both very short and long visual events (2 seconds to 10 minutes) in long videos.
Besides, post-processing is also applied to exclude static scenes and transition clips, ensuring each visual event contains rich and meaningful content.

\noindent \textbf{Generic semantics-aware audio event boundary.}
Although audio is crucial for temporal video understanding, no method exists for detecting generic audio event boundaries without pre-defined categories.
To fill this gap, we design a generic method that segments long audio sequences into semantically coherent clips leveraging distinct audio properties.
Specifically, we first extract audio features using Mel-Frequency Cepstral Coefficients (MFCC)~\cite{juvela2018speech,hansen2015speaker},
which captures the audio's key frequency characteristics aligned with human auditory perception.
Then, we compute the mean of MFCC deltas (first-order differences) to summarize temporal variations, identifying values above a threshold as significant audio transitions.
Since such splitting only considers changes in spectral properties and overlooks semantic transitions, we further adopt CLAP~\cite{wu2023large} to extract audio embeddings, stitching adjacent clips if they are semantically similar.
We also implement additional procedures to merge segments split by abrupt changes, \eg, pauses between spoken words or sudden shifts in music.  

\noindent \textbf{Omni-modal event boundary.}
After identifying the single-modality event boundaries, we found visual events tend to be longer than audio ones, with multiple audio events often occurring within a single visual event. 
To define omni-modal event boundaries, we primarily rely on visual boundaries while preserving the integrity of audio events. 
Specifically, for each visual event, we set its start time as the beginning of the omni-modal event and include all overlapping audio events without truncation to determine the event's end. 
By doing this, we can maximize the semantic integrity and coherence of events across both modalities.

\subsection{Omni-Modal Event Captioning}
\label{sec:3.3}
We develop a comprehensive relation-aware captioning strategy to generate high-quality omni-modality dense captions for long videos by integrating all modality information (\ie, visual, audio, and speech). 
As shown in Fig.~\ref{fig:dataset_pipeline}, we first generate detailed dense captions for each modality, and then integrate them to obtain omni-modality captions, emphasizing the reasoning of semantic and temporal relationships across events from various modalities.

\begin{figure*}[ht]
  \centering
  \setlength{\abovecaptionskip}{1.0mm}
   \includegraphics[width=1.0\linewidth]{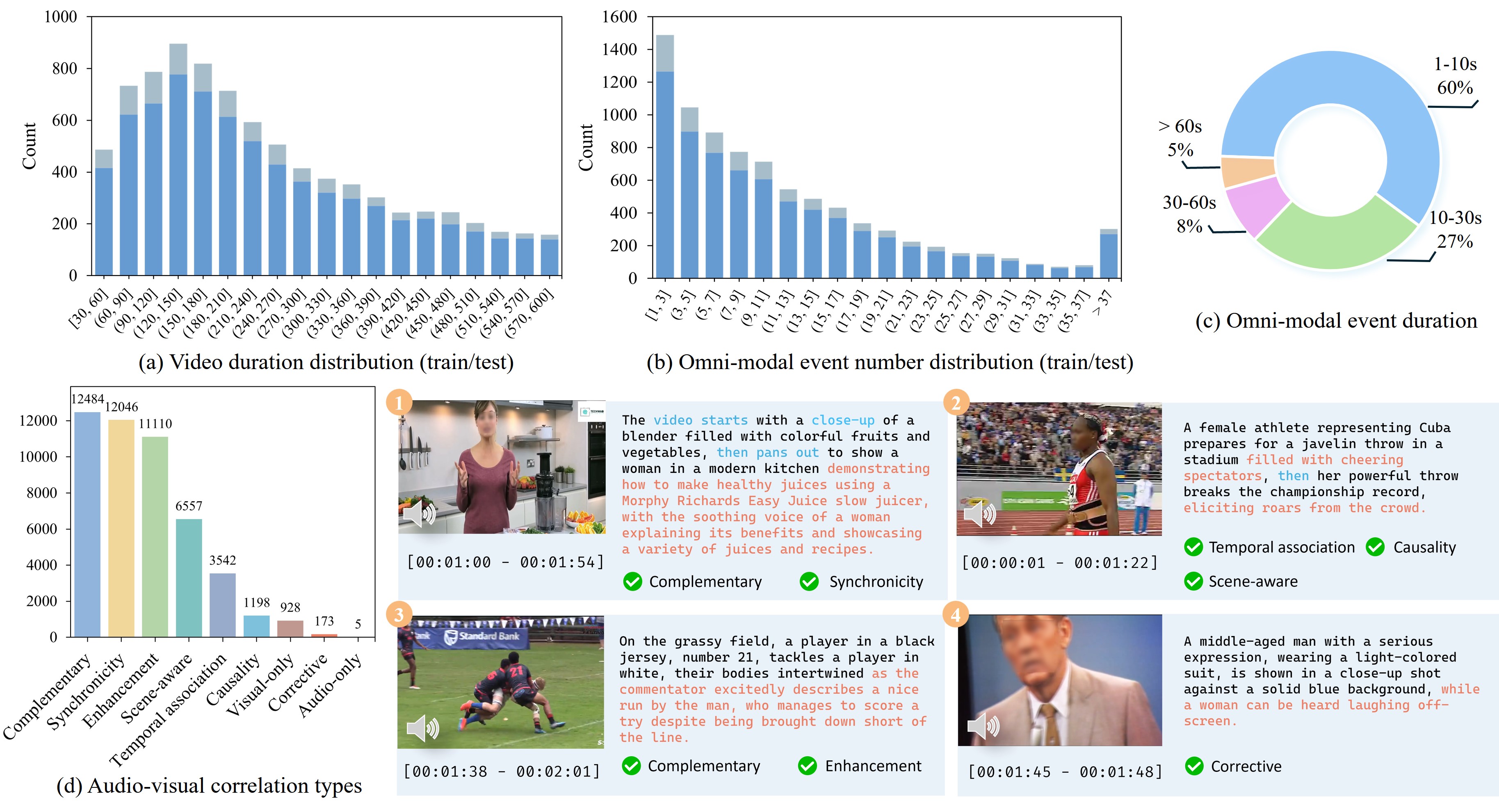}
   \vspace{-5mm}
   \caption{Statistics of LongVALE benchmark. (a) Video duration distribution of both training and test sets. (b) Distribution of the number of omni-modal events in videos for both training and test sets. (c) Distribution of omni-modal event duration. (d) Distribution of audio-visual correlation types. The examples of omni-modal events with different audio-visual correlations are also illustrated.}
   \label{fig:dataset_overall}
     \vspace{-5mm}
\end{figure*}

\noindent \textbf{Dual-focus visual captioning.}
Existing video automatic captioning strategies~\cite{rizve2024vidla,chen2023vast,chen2024sharegpt4video} only use image captioners like BLIP~\cite{li2022blip} and GPT-4V~\cite{Gpt-4v} to describe uniformly sampled frames, lacking the awareness of temporal dynamic knowledge in videos. 
In contrast, we focus on both spatial visual details and significant dynamic information (\ie, actions and camera movements), and also consider the complexities of long video events. 
Specifically, we employ LLaVA-NeXT-Video~\cite{zhang2024llavanextvideo} to caption each video event. 
However, the model's performance drops with longer clips, so we divide clips longer than 30 seconds and caption them separately to preserve as many dynamic details as possible.
Additionally, we sample the center frame of each cropped clip as the keyframe and use GPT-4o~\cite{Gpt-4o} to generate comprehensive image captions. 
Incorporating such precise spatial details (\ie, OCR, object appearance, and scene context) helps improve caption quality and effectively correct errors caused by hallucinations from the video model.

\noindent \textbf{Audio and speech captioning.}
To capture the full details for each audio event, we focus on both general audio and speech captioning, as they are equally essential for audio content understanding.
Specifically, we employ Qwen-Audio~\cite{chu2023qwen}, a general large audio-language model, to obtain detailed audio descriptions. 
At the same time, Whisper-Large~\cite{radford2023robust}, a strong automatic speech recognition (ASR) model, is applied to get accurate subtitles if the clip contains speech content.  
Additionally, we perform further data refinement to minimize the impact of hallucinations.

\noindent \textbf{Relation-aware omni-modal event captioning.}
Our goal is to generate high-quality omni-modal event captions by reasoning about audio-visual correlations and temporal dynamics across modality-specific events.
Specifically, instead of simply concatenating modality-specific captions~\cite{chen2023vast,liu2024valor}, we instruct Gemini-1.5-Pro~\cite{Gemini} to establish meaningful connections between them, such as analyzing whether audio events are visible, identifying sound sources, and reasoning about causality, \etc.
Additionally, we provide the model with single-modal event boundaries, guiding it to perceive fine-grained temporal changes within an omni-modal event, such as camera movements or sequential speaking in conversations, and summarize these to create fine-grained time-aware descriptions.  
Moreover, we feed the generated captions of previous omni-modal events as context, enabling the model to produce a more coherent and accurate caption for the current omni-modal event, thereby ensuring a more seamless dense narrative for a long video.

\subsection{Subset Split and Manual Efforts}
\vspace{-1mm}
Through the above carefully designed scalable pipeline, we automatically collect 8,411 high-quality long videos with highly valuable dense omni-modal event annotations. 
We then meticulously divide the data into training and test sets, ensuring consistent data distribution (\eg, video duration and event counts) between them.
The final training and test sets consist of 7,240 and 1,171 long videos, respectively.

\noindent\textbf{Test set manual check and correction.}
To build a high-quality test set, we conducted thorough manual checks and corrections.
We developed an interface where one group reviews entire videos to verify the accuracy of omni-modal event boundaries and captions. Another group then corrects flagged errors to ensure high precision of annotations.

\subsection{Statistic Analysis}
\label{sec:3.5}
\vspace{-1mm}
Overall, our LongVALE is the first-ever omni-modality long video understanding benchmark with dense event-level annotations. The statistics are shown in Fig.~\ref{fig:dataset_overall}.
It includes 8,411 long videos spanning over 549 hours, with an average video duration of 3.9 minutes. 
Notably, the dataset contains 105,730 omni-modal events (91,863/13,867 in train/test split), each annotated with accurate temporal boundaries and omni-modality relation-aware captions.  

\noindent \textbf{Omni-modal event distribution.}
As shown in Fig.~\ref{fig:dataset_overall}(b), a large number of videos contain multiple omni-modal events, with an average of 12.6 events per video. 
Besides, the events have various lengths spanning from 1 second to even 10 minutes, with an average length of 16.7 seconds. 
Most events are relatively short, with 60\% lasting under 10 seconds and 97\% under 30 seconds as shown in Fig.~\ref{fig:dataset_overall}(c). 
Moreover, all events are non-overlapping and cover 89\% of the total video duration, highlighting the dense nature and complexity of real-world multi-modal long videos.

\noindent \textbf{Omni-modal caption characteristics.}
We highlight that omni-modal captions exhibit unique characteristics of \textit{audio-visual correlations} and \textit{fine-grained temporal dynamics}.
We further analyze all omni-modal captions in our test set using Gemini-1.5-Pro~\cite{Gemini} to identify these characteristics:
\textbf{1)} We define seven types of audio-visual correlations with two unimodal types (\ie, visual/audio-only). 
The occurrences of each type are shown in Fig.~\ref{fig:dataset_overall}(d).
It indicates that \textit{complementary}, \textit{synchronicity} and \textit{enhancement} are common correlation types. 
As shown in Fig.~\ref{fig:dataset_overall}, the speech content \textit{complements} the visual scene, where the visible woman speaker reflects \textit{synchronized} visual and audio events (example 1), and the off-screen commentator's excited tone \textit{enhances} the intense competition atmosphere (example 3).
Besides, there also exist other correlations, such as in the second example, where the crowd's cheers indicate a \textit{scene-aware} context of the stadium and the athlete's success triggers roars, demonstrating the \textit{causality} and \textit{temporal association} between the visual and audio events. 
Moreover, audio can provide \textit{corrective} insights for misleading visual cues, as seen in the fourth example where background laughter reveals a comedy show despite the man's serious face.
\textbf{2)} Additionally, the quantitative analysis also indicates that 78\% of the omni-modal captions capture fine-grained temporal dynamics. As highlighted in the blue words in Fig.~\ref{fig:dataset_overall}, the captions reflect a high-level understanding of sequential sub-events across different modalities, \eg, camera zooms and plot progressions. 
It underscores the extensive and complex semantic information embedded in our omni-modal captions.

\section{LongVALE-LLM}
Inspired by the advancements in video large language models (video LLMs) on diverse video tasks~\cite{li2023videochat,zhang2023video,huang2024vtimellm}, we present LongVALE-LLM, a video LLM designed to handle omni-modal event understanding in long videos.

\subsection{Overall Architecture}
\begin{figure}[t]
  \centering
  \setlength{\abovecaptionskip}{1.0mm}
   \includegraphics[width=1.0\linewidth]{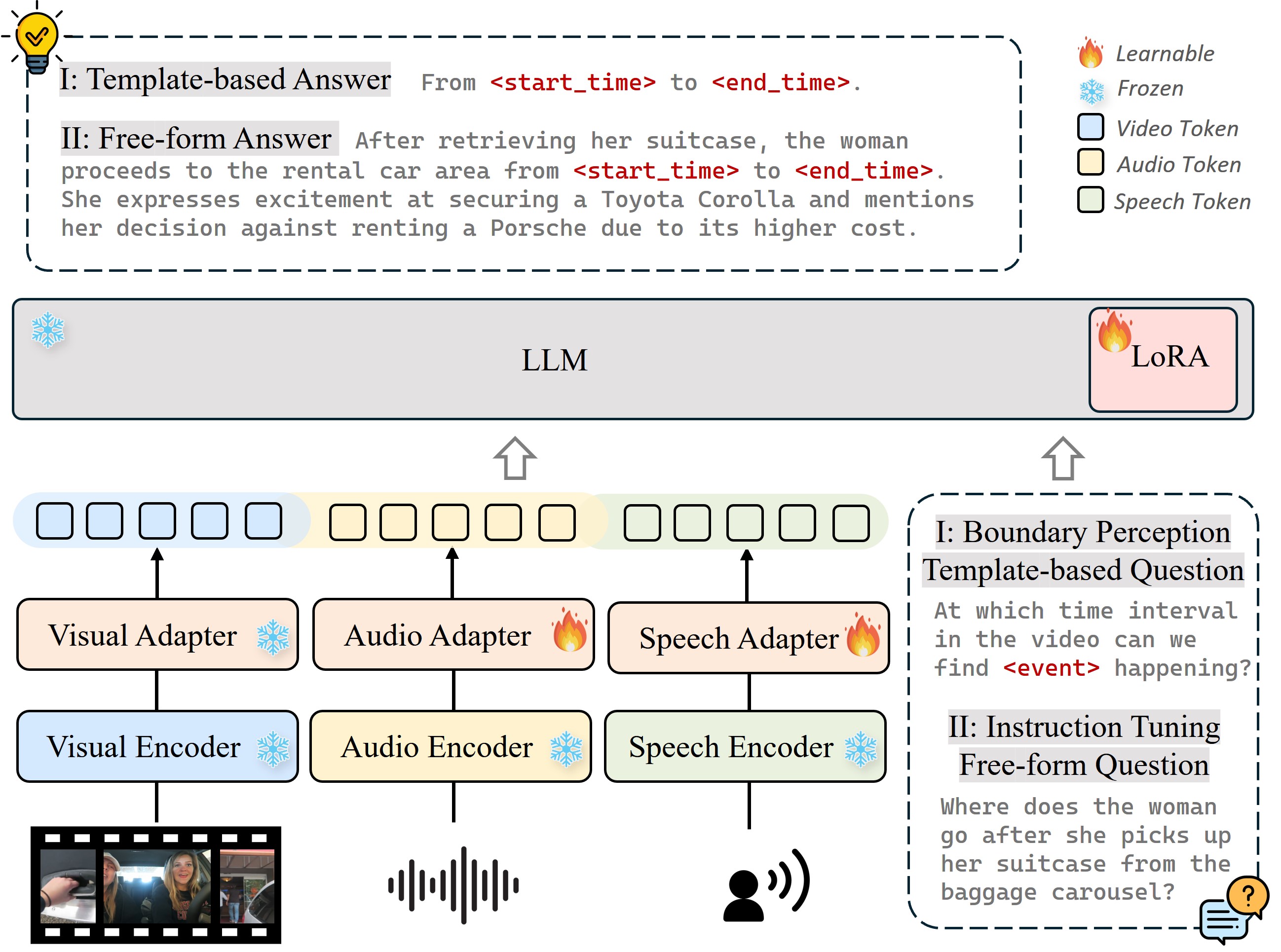}
   \caption{LongVALE-LLM architecture with boundary perception and instruction tuning stages using our LongVALE dataset.}
   \label{fig:baseline}
     \vspace{-6mm}
\end{figure}

Figure~\ref{fig:baseline} illustrates the overall architecture of LongVALE-LLM. Given a long video, the multi-modal encoders first extract modality-specific token features, which are then mapped into the LLM's embedding space via the multi-modal adapters. The embeddings from different modalities are concatenated along the sequence dimension and combined with the task instruction to form the prefix input to the LLM. The LLM is trained with an auto-regressive objective to generate responses that align with both the instruction and the omni-modality content.

\subsection{Training Recipe}
Our LongVALE, as the first-ever omni-modality long video benchmark with fine-grained annotations, serves as a highly valuable data source for training a Video-LLM capable of both cross-modal reasoning and fine-grained temporal understanding. Extending the boundary-aware training strategy from visual-only Video-LLMs~\cite{huang2024vtimellm}, we introduce omni-modal boundary perception and instruction tuning to allow omni-modal event understanding. Note that in both training stages, we train the audio and speech adapters, and the LLM (using LoRA~\cite{hu2021lora}), while keeping the visual adapter frozen, which is pre-trained with LCS-558k dataset~\cite{liu2024improved}.  

\noindent\textbf{Omni-modal boundary perception tuning.} The training stage focuses on enabling the LLM to comprehend omni-modal events within a video and align them with their corresponding temporal boundaries.  
For training data, we transform the omni-modal event annotations of each video into template-based dialogue data suitable for training LLM. The dialogues include both single-turn and multi-turn QA dialogues similar in VTimeLLM~\cite{huang2024vtimellm}. Single-turn QA tasks focus on omni-modal dense video captioning, while multi-turn QA tasks handle omni-modal video grounding and segment captioning. 
We only generate one set of dialogues for each video, yielding 7,240 QA dialogues. Additionally, we also add visual-only data~\cite{huang2024vtimellm} in this tuning stage.

\begin{table*}
  \centering
  \resizebox{0.9\linewidth}{!}{ 
  \begin{tabular}{l|cc|cccc|ccc|cccc}
    \toprule
    \multirow{2}{*}{Model} & \multirow{2}{*}{A\&V}& \multirow{2}{*}{TU} & \multicolumn{4}{c}{Omni-TVG} & \multicolumn{3}{c}{Omni-DVC} & \multicolumn{4}{c}{Omni-SC}  \\
    \cmidrule(lr){4-14} 
     & && R@0.3 & R@0.5 & R@0.7 & mIoU & S & C & M & B & R & C & M \\
     \midrule
    VideoChat (7B)~\cite{li2023videochat} & \ding{55} & \ding{55} & 2.2 & 0.9 & 0.4 & 3.0 & 0.7 & 0.2 & 0.9 & 0.5 & 9.6 & 0.0 & 8.2  \\
    VideoChatGPT (7B)~\cite{maaz2023video} & \ding{55} & \ding{55} & 4.9 & 2.0 & 0.9& 5.0 & 0.7& 0.1& 0.9&0.4& 14.0& 0.9 & 5.9 \\
    VideoLLaMA (7B)~\cite{zhang2023video}& \checkmark & \ding{55}  & 2.5 & 1.1 & 0.3 & 1.9 & 0.6& \underline{0.6}& 0.9& 0.9& 11.5& 0.1& 8.9 \\
    PandaGPT (7B)~\cite{su2023pandagpt} & \checkmark & \ding{55} &2.5& 1.0& 0.3 & 2.2 & 0.5 & 0.0 & 0.6 & 0.6& 14.9 & 0.3 & 8.9 \\
    NExT-GPT (7B)~\cite{wu2023next} & \checkmark & \ding{55} & 4.3 & 1.9 & 0.7 & 4.0 &0.2 & 0.1& 0.3 & 0.4 & 10.2& 0.0 & 8.1 \\
    TimeChat (7B)~\cite{ren2024timechat} & \ding{55} & \checkmark & 5.8 & 2.6 & 1.1 & 5.2 & 1.6 & 0.1 & 1.4 & \underline{1.2} & \underline{16.1} & \underline{1.6} & \underline{10.0} \\
    VTimeLLM (7B)~\cite{huang2024vtimellm} & \ding{55} & \checkmark & \underline{7.5} & \underline{3.4} & \underline{1.3} & \underline{6.4} & \underline{2.4} & 0.2& \underline{2.0} &1.0&14.5&1.6&5.5  \\
    \midrule
    \rowcolor{gray!20} LongVALE-LLM (7B) (ours) & \checkmark & \checkmark & \textbf{15.7}&\textbf{8.6}&\textbf{3.9}&\textbf{11.0}&\textbf{2.8}&\textbf{7.9}&\textbf{4.7}&\textbf{5.6}&\textbf{22.4}&\textbf{20.3}&\textbf{10.9}\\
    \bottomrule
  \end{tabular}}
  \vspace{-1.5mm}
  \caption{Comparison with existing Video LLMs for omni-modal temporal video grounding (Omni-TVG), dense video captioning (Omni-DVC), and segment captioning (Omni-SC) tasks on our LongVALE test set. A\&V: support both video and audio input. TU: support fine-grained temporal understanding. S: SODA\_c. C: CIDEr. M: METEOR. B: BLUE-4. R: ROUGE-L.}
  \label{tab:SOTA}
  \vspace{-5mm}
\end{table*}

\noindent\textbf{Omni-modal instruction tuning.}
Although our model demonstrates the ability to perceive omni-modal event boundaries after boundary perception tuning, its outputs tend to overfit to templated answers. 
To improve the model's ability to follow human instructions for more comprehensive omni-modal event reasoning, we create high-quality instruction-tuning data based on our LongVALE.
We convert all video annotations into high-quality QA dialogues using Gemini-1.5-Pro~\cite{Gemini}. For each video, we prompt the LLM to analyze omni-modal event boundaries and captions, generating free-form dialogues that emphasize temporal perception and reasoning, which may encompass a variety of tasks. {\color{black}The prompt can be found in Appendix.}
As a result, we generate an omni-modal instruction dataset containing 25.4K high-quality QA dialogues with an average of 3.6 distinct dialogues per video.
Besides, we also incorporate extra visual-only instruction data~\cite{huang2024vtimellm} to further enhance the model's descriptive capabilities.

\section{Experiments}

\subsection{Experiment Setup}
\noindent\textbf{Implementation details.}
We use CLIP VIT-L/14~\cite{radford2021learning} as the visual encoder. We uniformly sample 100 frames and encode each frame individually, representing each by the feature of the CLS token for efficiency. BEATs~\cite{chen2022beats} is employed as the audio encoder, and Whisper-Large-v2~\cite{radford2023robust} is used for speech encoding. Both audio and speech are processed from the waveforms of 5.12-second clips, with the number of tokens varying according to the video's duration. 
The audio and speech adapters are just randomly initialized. Furthermore, Vicuna-v1.5-7b~\cite{chiang2023vicuna} is adopted as the large language model. {\color{black} More details are in Appendix}.

\noindent\textbf{Evaluation metrics.}
Using our LongVALE test set, we evaluate models on three omni-modal fine-grained understanding tasks.
For omni-modal temporal video grounding, we report Recall@1 at IoU thresholds of \{0.3,0.5,0.7\} and mean IoU (mIoU).
For omni-modal dense video captioning, we assess caption quality using CIDEr~\cite{vedantam2015cider} and METEOR~\cite{banerjee2005meteor}, and employ SODA\_c~\cite{fujita2020soda} for overall story-level evaluation.
For omni-modal segment captioning, we use BLEU-4~\cite{papineni2002bleu}, ROGUE-L~\cite{lin2004automatic}, METEOR~\cite{banerjee2005meteor} and CIDEr~\cite{vedantam2015cider} for standard caption quality evaluation.

\subsection{Main Results}
Table~\ref{tab:SOTA} presents the comparison results of our model with existing open-sourced video LLMs on the three omni-modal fine-grained tasks on our LongVALE test set, where we ensure optimal evaluation of the existing models.
Our LongVALE-LLM (7B) supports video, audio and speech input with fine-grained temporal understanding ability, and outperforms other video LLMs by a significant margin across all three tasks.
Despite VideoLLaMA~\cite{zhang2023video}, PandaGPT~\cite{su2023pandagpt} and NExT-GPT~\cite{wu2023next} also support audio-visual input, they are limited to processing a few video frames (\eg, 8 frames), resulting in poor performances on fine-grained, time-sensitive tasks. 
Besides, VTimeLLM~\cite{huang2024vtimellm} and TimeChat~\cite{ren2024timechat} can understand specific temporal events in videos, but they focus solely on visual events and fail to incorporate crucial audio information for a complete video understanding. 
Therefore, it is essential to integrate both audio and visual information with boundary-aware training on rich omni-modality data, \ie, LongVALE, to achieve precise video comprehension.

\begin{table}
  \centering
  \resizebox{0.9\linewidth}{!}{ 
  \begin{tabular}{l|c|cc}
    \toprule
    Method & \#Pairs & AVSD & Music-AVQA \\
    \midrule
    PandaGPT (13B)~\cite{su2023pandagpt} & 128M & 26.1 & 33.7 \\
    Macaw-LLM (7B)~\cite{lyu2023macaw} & 0.3M & 34.3 & 31.8 \\
    VideoLLaMA (7B)~\cite{zhang2023video} & 2.8M & 36.7 & 36.6 \\
    X-InstructBLIP (13B)~\cite{panagopoulou2023x} & 32M & - & 44.5 \\
    AV-LLM (13B)~\cite{shu2023audio} & 1.6M & 52.6 & 45.2 \\
    OneLLM (7B)~\cite{han2024onellm} & 1007M & - & 47.6 \\
    AVicuna (7B)~\cite{tang2024avicuna} & 1.1M & \underline{53.1} & \textbf{49.6} \\
    \midrule
    \rowcolor{gray!20} LongVALE-LLM (7B) (ours) & 0.7M & \textbf{54.8} & \underline{49.4}\\
    \bottomrule
  \end{tabular}}
  \vspace{-1.5mm}
  \caption{Comparison with existing LLM-based methods on open-ended audio-visual question answering benchmarks on a zero-shot setting. \# Pairs: the adopted instruction-response pairs.}
  \label{tab:AVQA}
  \vspace{-5mm}
\end{table}

\subsection{Zero-Shot Performance on General AVQA}
Besides the ability for fine-grained omni-modality video understanding tasks, we also explore whether our model trained on LongVALE can address a broader range of audio-visual questions in a zero-shot setting. 
We employ the AVSD~\cite{alamri2019audio} and Music-AVQA~\cite{li2022learning} benchmarks and conduct a GPT-assisted evaluation to assess the accuracy of the generated answers as same as the protocol~\cite{maaz2023video} used by other LLM-based methods shown in Tab.~\ref{tab:AVQA}.
We can observe that, despite using significantly less data, our model surprisingly achieves state-of-the-art performance on AVSD and highly competitive results on Music-AVQA.  
Notably, existing multi-modal LLMs all rely on large amounts of audio-related training data. 
For example, OneLLM~\cite{han2024onellm} and AVicuna~\cite{tang2024avicuna} use 460K and 350K audio/audio-visual instruction-response pairs for training, respectively. 
In contrast, we achieve even better results using only total 32.7K audio-visual samples from our LongVALE dataset, accounting for less than 10\% of the data they use. 
This clearly validates the robustness and generalization of our LongVALE-trained model, demonstrating that the comprehensive omni-modality captions in our dataset can effectively enhance the model's general cross-modal reasoning capability.

\begin{table}
  \centering
  \resizebox{\linewidth}{!}{ 
  \begin{tabular}{c|c|c|c|c}
    \toprule
    BP & IT &Omni-TVG$_{mIoU}$ & Omni-DVC$_{CIDEr}$ & Omni-SC$_{CIDEr}$  \\
     \midrule
     V~\cite{huang2024vtimellm} & \ding{55} &12.6 / 3.1 & 0.1 / 0.1 & 0.4 / 0.3  \\
     V~\cite{huang2024vtimellm} + Ours & \ding{55} & 25.6 / 7.0 & 7.3 / 7.0  & 21.1 / 17.5 \\
     V~\cite{huang2024vtimellm} + Ours & V~\cite{huang2024vtimellm} & 11.7 / 3.8& 0.2 / 0.2 & 3.1 / 2.4  \\
     \rowcolor{gray!20} V~\cite{huang2024vtimellm} + Ours & V~\cite{huang2024vtimellm} + Ours & 26.0 / 7.3 &  7.8 / 7.7 & 25.1 / 19.4 \\
    \bottomrule
  \end{tabular}}
  \vspace{-1.5mm}
  \caption{Ablation study of the data used in different training stages. BP: boundary perception. IT: instruction tuning. V~\cite{huang2024vtimellm}: the visual-only data used in VTimeLLM~\cite{huang2024vtimellm}. 
  }
  \label{tab:ablation_stage}
  \vspace{-2mm}
\end{table}

\begin{table}
  \centering
  \resizebox{\linewidth}{!}{ 
  \begin{tabular}{c|c|c|c}
    \toprule
     Audio-Visual Correlation &Omni-TVG$_{mIoU}$ & Omni-DVC$_{CIDEr}$ & Omni-SC$_{CIDEr}$  \\
     \midrule
    \ding{55} & 23.7 / 6.5 & 3.5 / 3.4 & 10.4 / 10.2 \\
    \rowcolor{gray!20} \checkmark & 25.6 / 7.0 & 7.3 / 7.0  & 21.1 / 17.5 \\
    \bottomrule
  \end{tabular}}
  \vspace{-1.5mm}
  \caption{Ablation study of audio-visual correlation reasoning in captioning for the boundary perception training stage.}
  \label{tab:ablation_avc}
  \vspace{-2mm}
\end{table}

\begin{table}
  \centering
  \resizebox{0.9\linewidth}{!}{ 
  \begin{tabular}{c|c|c|c}
    \toprule
     Modality &Omni-TVG$_{mIoU}$ & Omni-DVC$_{CIDEr}$ & Omni-SC$_{CIDEr}$ \\
    \midrule
     V & 12.6 / 2.2 & 5.9 / 3.6 &14.6 / 11.7 \\
     V+A & 15.6 / 3.1 & 7.2 / 5.0 & 17.6 / 14.7 \\
     V+S & 15.2 / 2.9 & 6.7 / 5.2 & 17.3 / 14.6  \\
     \rowcolor{gray!20} V+A+S & 17.1 / 3.0 & 7.8 / 5.6 & 18.9 / 15.4 \\
    \bottomrule
  \end{tabular}}
  \vspace{-1.5mm}
  \caption{Ablation study of models trained on LongVALE with different modalities. V: vision. A: audio. S: speech. }
  \label{tab:ablation_modality}
  \vspace{-5mm}
\end{table}

\subsection{Ablation Study and Qualitative Results}
We provide detailed ablation studies about our training data, training stages, and modalities as follows.
Since LongVALE is a challenging benchmark, we split the test set into easy and hard subsets based on the ratio of average event duration to video length for more in-depth analysis. 
The easy subset ratios from 100\% to 9.3\% (585 videos), and the hard subset ratios from 9.3\% to 0.95\% (586 videos). 
Note that in Tab.~\ref{tab:ablation_stage}-\ref{tab:ablation_modality}, easy/hard subset results are on the left/right.

\noindent\textbf{Impact of LongVALE in different training stages.}
In Tab.~\ref{tab:ablation_stage}, we observe significant performance boosts across all tasks by adding our data in both boundary perception and instruction tuning stages. 
It demonstrates the essential role of our data in promoting omni-modality fine-grained video understanding.
Besides, the necessity of instruction tuning is evident when comparing the second and last rows. 
It shows consistent improvements across all three tasks, especially in segment captioning, indicating instruction tuning on high-quality dialog data further enhances comprehension and reasoning abilities for omni-modal events.

\noindent\textbf{Importance of audio-visual correlation (AVC).}
We simply concatenate the generated single-modality captions in Sec.~\ref{sec:3.3} as a naive caption for each omni-modal event boundary without AVC reasoning.
Then, we convert them into template-based dialogues for boundary perception tuning. 
The results in Tab.~\ref{tab:ablation_avc} show that the model trained with our omni-modality captions with AVC reasoning achieves significantly better performance, especially for captioning tasks, demonstrating the effectiveness of AVC to facilitate the model's capability of cross-modal reasoning.

\noindent\textbf{Impact of using different modalities.}
In Tab.~\ref{tab:ablation_modality}, we can see that adding audio or speech modality significantly improves the performances across all three tasks, with the best results achieved when all three modalities are used. 
This highlights the strength of omni-modality input for multi-modal video understanding.
Note that we use only our dataset during both training stages to provide a clearer comparison. 

\begin{figure}[t]
  \centering
  \setlength{\abovecaptionskip}{1.0mm}
   \includegraphics[width=1.0\linewidth]{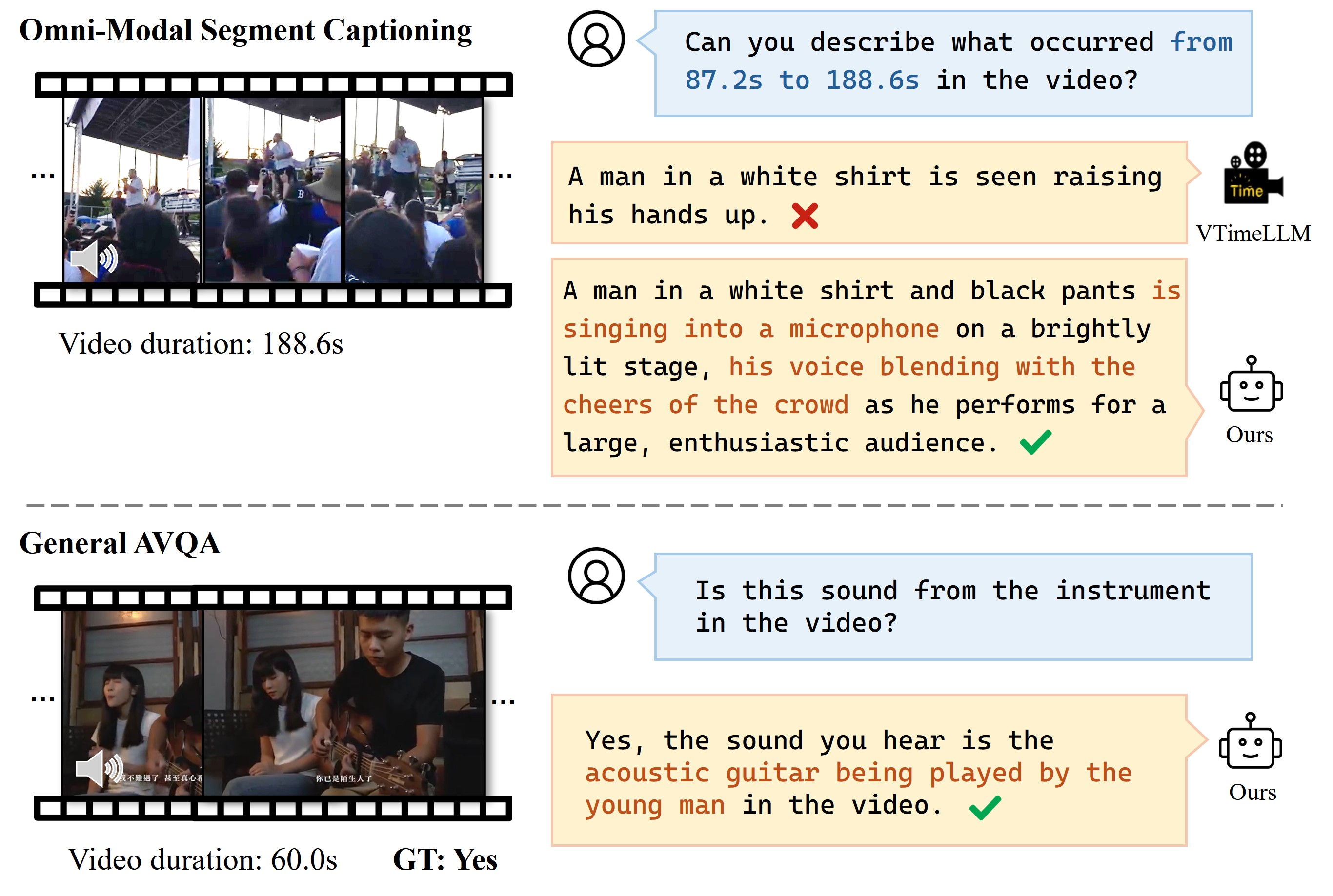}
   \caption{Qualitative results. The orange text highlights audio-visual correlation for accurate and complete video understanding. Samples are from LongVALE and Music-AVQA test sets. 
   }
   \label{fig:visualization}
     \vspace{-5mm}
\end{figure}

\noindent\textbf{Qualitative results.}
In Fig.~\ref{fig:visualization}, VTimeLLM~\cite{huang2024vtimellm} with only visual data, misidentifies singing as raising hands. In contrast, our model integrates audio signals (\ie, man singing, crowd cheers) and performs cross-modal reasoning to accurately and comprehensively describe the event in the specific moment. Besides, our model effectively associates audio-visual information to provide correct answers for general AVQA. {\color{black}More examples can be found in Appendix.}

\section{Conclusion}
We present a scalable pipeline to build LongVALE, the first benchmark for omni-modality fine-grained video understanding, featuring 105K omni-modal events with temporal boundaries and relation-aware captions. 
Benefiting from the dataset, our model exhibits distinct capabilities of both cross-modal reasoning and fine-grained temporal understanding that are absent in existing video LLMs, making a crucial step toward an intelligent video assistant.
In the future, we will expand our LongVALE with more high-quality data and advance the model's architecture to improve video semantic density and cross-modal interaction.

{
    \small
    \bibliographystyle{ieeenat_fullname}
    \bibliography{main}
}

\clearpage
\maketitlesupplementary

\section{More Details of LongVALE Benchmark}
\subsection{Quantitative Analysis of Event Boundaries}
\label{sec:boundary}
To quantitatively verify the semantic coherence of segmented events of different modalities, we introduce Max Running Semantic Difference (MRSD), inspired by~\cite{chen2024panda}. For a $n$-second event clip, we compute the embedding for each second as $\{f_{1}, \dots, f_{n}\}$, and get the most significant semantic change within the clip, denoted as:   
\begin{equation}
    \mathrm{max}(\{\mathrm{Diff}(f_{i},f_{i+1}) | i \in [1, n-1]\}).
\end{equation}
We apply ImageBind~\cite{girdhar2023imagebind} and CLAP~\cite{wu2023large} to extract embeddings for visual and audio clips, respectively. 
As in Tab.~\ref{tab:boundary}, for single-modal events, the clips after the second stitching stage effectively avoid being overly fragmentary while maintaining strong semantic coherence.
Further, although semantic shifts may occur between single-modal events within an omni-modal event, no event is truncated, ensuring the semantic integrity of all events from various modalities.

\begin{table}[ht]
  \centering
  \resizebox{\linewidth}{!}{ 
  \begin{tabular}{lccc}
    \toprule
    Method & MRSD-V$\downarrow$ & MRSD-A$\downarrow$ & Avg.len  \\
    \midrule
    Visual event boundary (splitting) & 0.531 & - & 3.0s\\
    Visual event boundary (stitching) & 0.532 & - & 10.7s \\
    \midrule
    Audio event boundary (splitting) & - & 0.676 & 1.5s\\
    Audio event boundary (stitching) & - & 0.703 & 5.8s \\
    \midrule
    Omni-modal event boundary & 0.601 & 0.784 & 16.7s \\
    \bottomrule
  \end{tabular}}
  \caption{Semantic coherence and event length analysis. We randomly sample 1K long videos in our LongVALE. }
  \label{tab:boundary}
\end{table}

\subsection{More Statistics}
\begin{figure}[ht]
  \centering
  \setlength{\abovecaptionskip}{1.0mm}
   \includegraphics[width=1.0\linewidth]{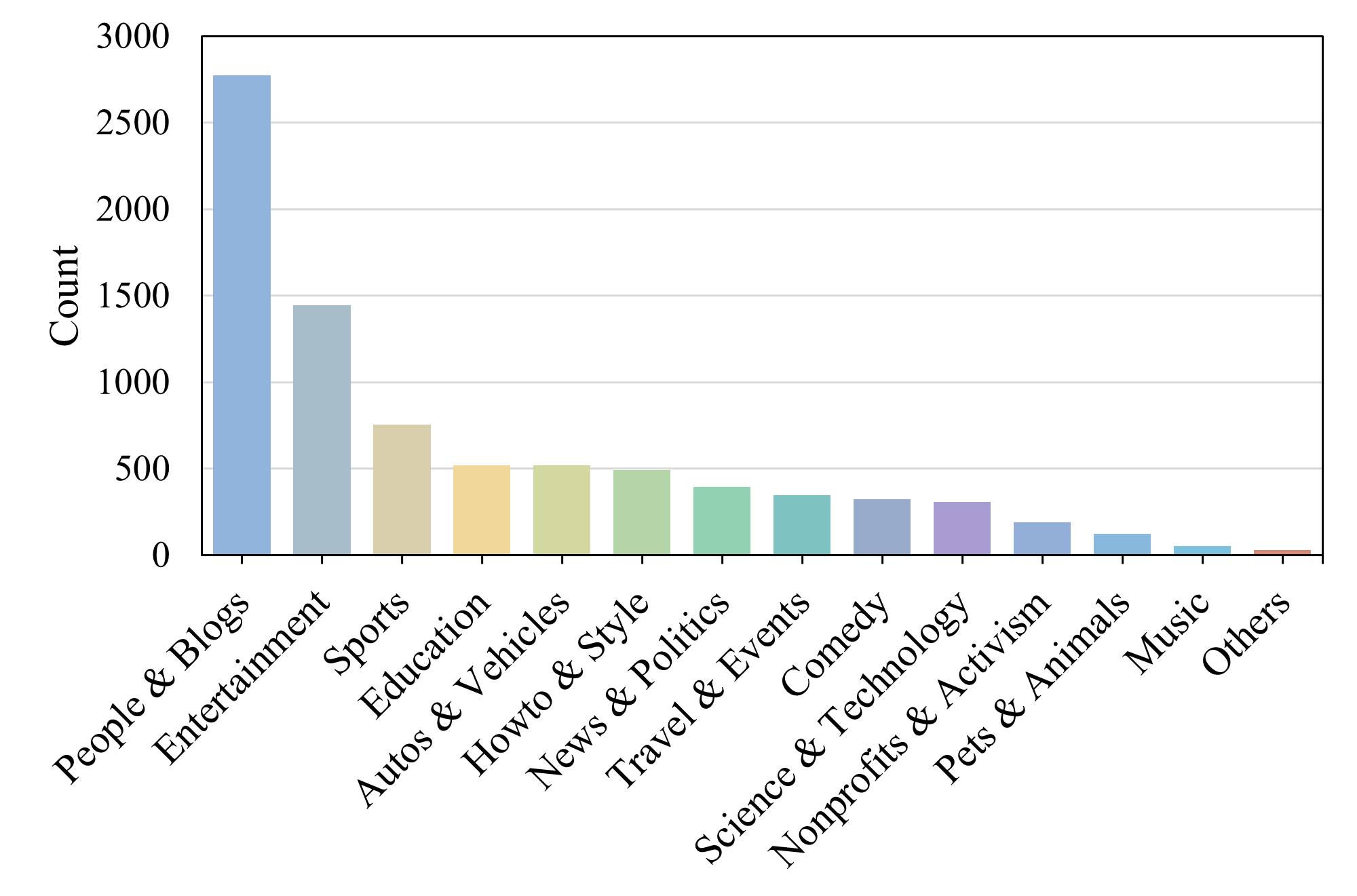}
   \caption{Distribution of video categories of LongVALE dataset.}
   \label{fig:category}
     \vspace{-4mm}
\end{figure}

\begin{figure*}[ht]
  \centering
  \setlength{\abovecaptionskip}{1.0mm}
   \includegraphics[width=1.0\linewidth]{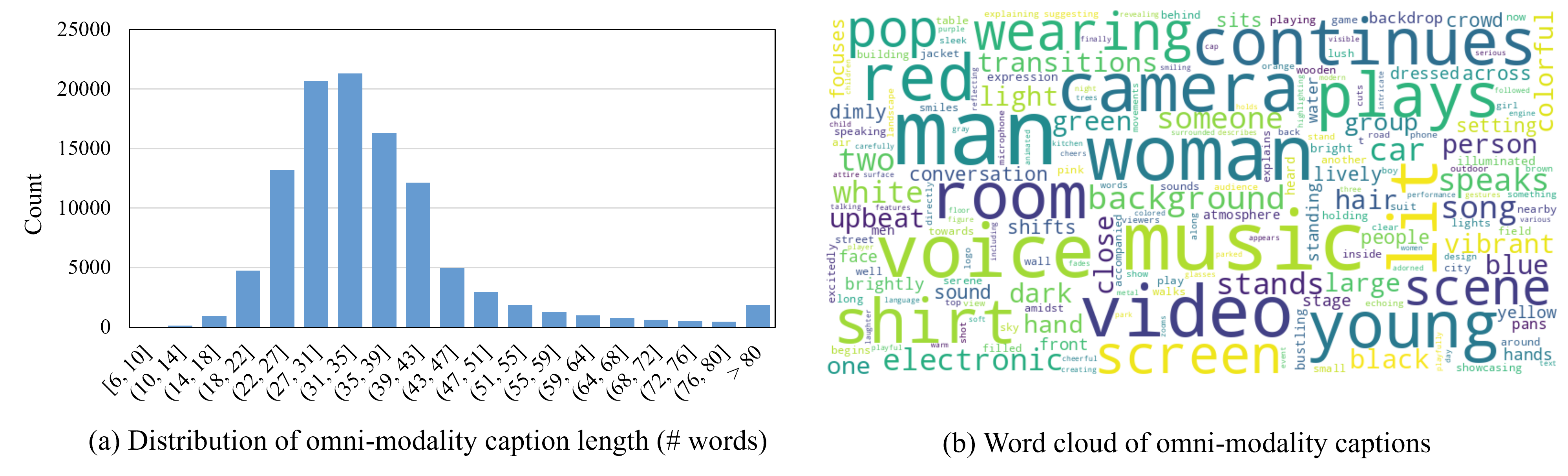}
   \caption{Distribution of omni-modality caption length and word cloud.}
   \label{fig:word}
     \vspace{+4mm}
\end{figure*}

Based on YouTube metadata, we further analyze the distribution of video categories, as shown in Fig.~\ref{fig:category}. It reflects that our LongVALE covers a wide range of video topics. Besides, since our focus is on long-form videos with rich, event-driven storylines, the diversity of their content cannot be easily summarized by just a few simple categories.
Moreover, as shown in Fig.~\ref{fig:word}, we also illustrate the distribution of the lengths of our omni-modal event captions and visualize their word cloud to highlight the rich omni-modality content within the captions.

\subsection{Manual Check and Correction}
\begin{figure*}[ht]
  \centering
  \setlength{\abovecaptionskip}{1.0mm}
   \includegraphics[width=1.0\linewidth]{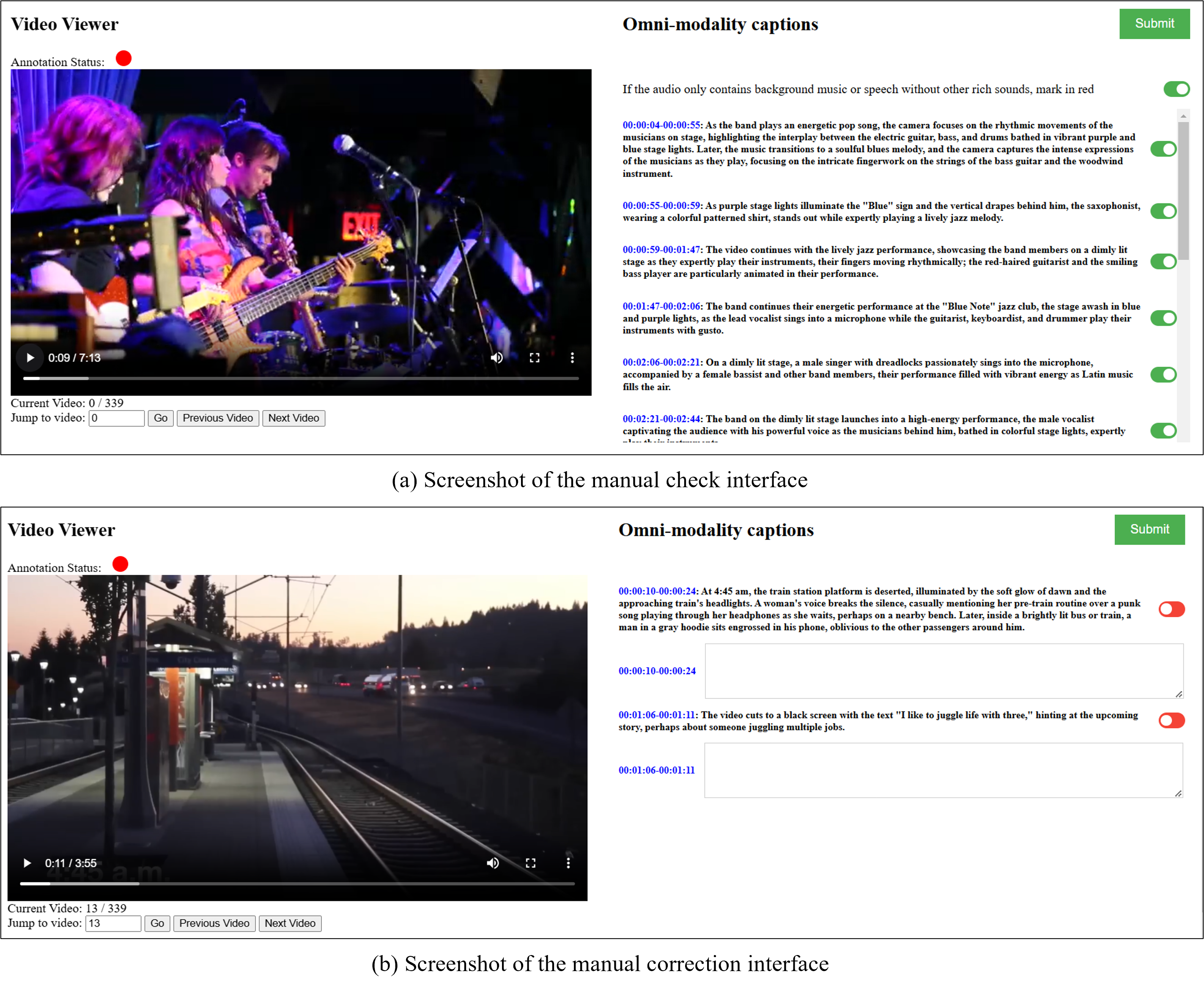}
   \caption{Screenshots of our manual check and correction interfaces.}
   \label{fig:manual}
\end{figure*}

During the manual check process, annotators are asked to check each omni-modal event and verify whether the caption and the corresponding temporal boundaries are accurate. Besides, videos containing only monotonous background music and speech are filtered out to ensure the dataset includes rich sound types.
Afterward, during the manual correction process, another group of annotators correct all inaccurate annotations and submit the revised versions. 
Totally, we checked 2K videos with each taking 3 minutes, and corrected about 300 errors, totally 115 human hours.
We show the interfaces in Fig.~\ref{fig:manual}

\subsection{Captioning and AV correlation Prompts}
\begin{figure*}[ht]
  \centering
  \setlength{\abovecaptionskip}{1.0mm}
   \includegraphics[width=1.0\linewidth]{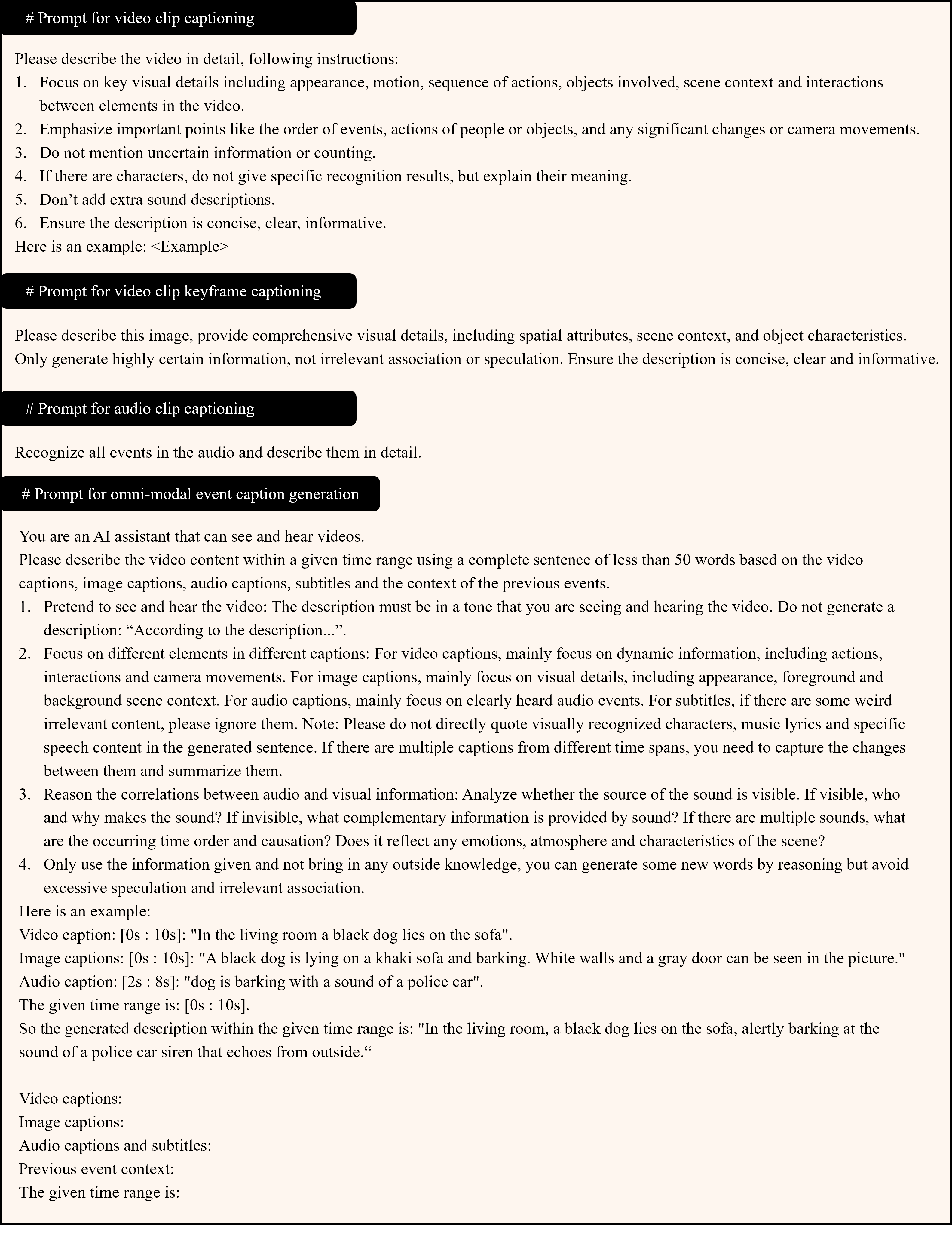}
   \caption{The prompts for the captioning of video clips, keyframes and audio clips, and integrating them for omni-modal events captions.}
   \label{fig:video_cap}
\end{figure*}

\begin{figure*}[h]
  \centering
  \setlength{\abovecaptionskip}{1.0mm}
   \includegraphics[width=1.0\linewidth]{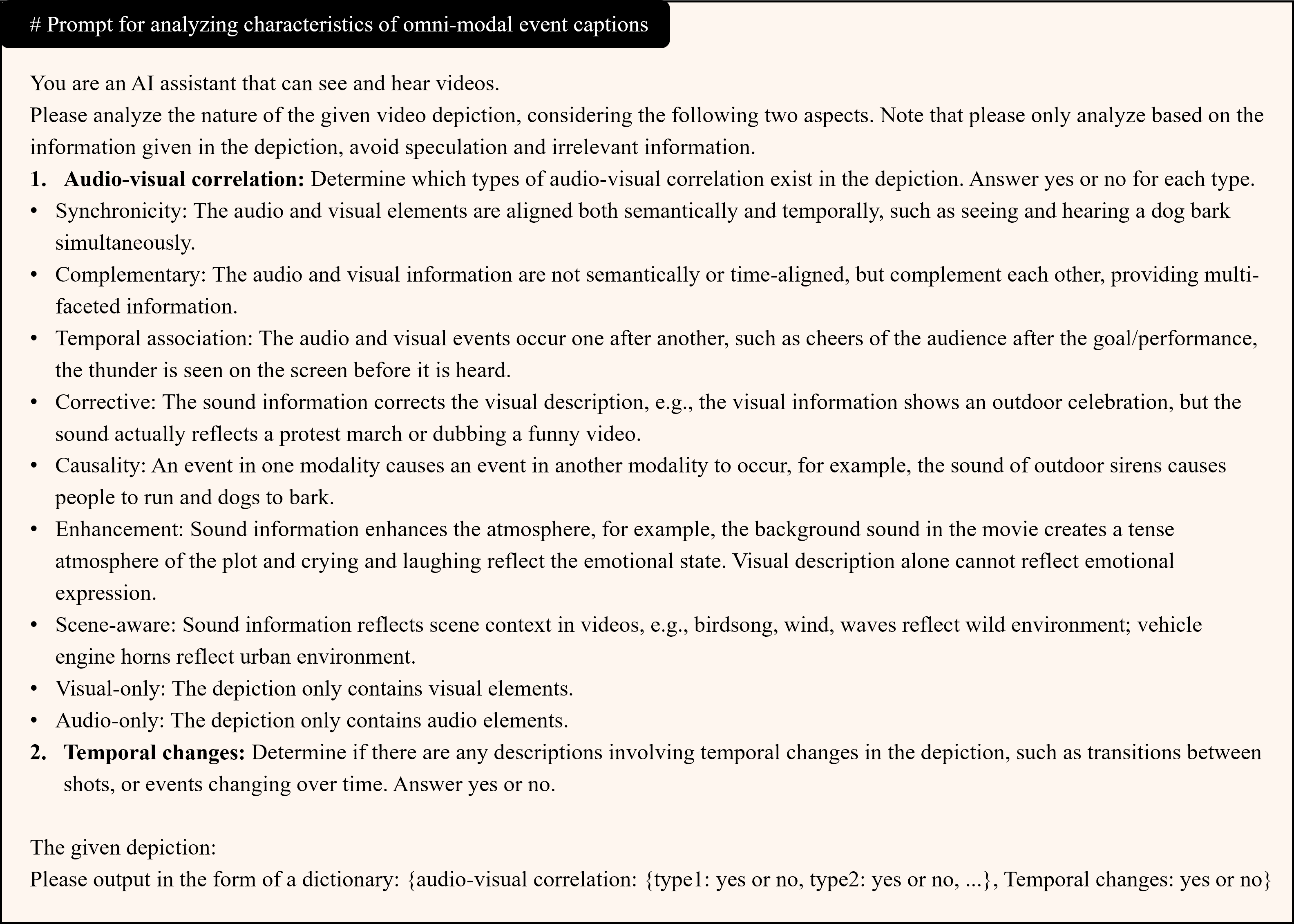}
   \caption{The prompt used to analyze and identify audio-visual correlations and temporal dynamics in our omni-modal event captions.}
   \label{fig:avc}
\end{figure*}

In Sec.{\color{blue}3.3}, for each segmented video clip, we apply LLaVA-NeXT-Video (34B)~\cite{zhang2024llavanextvideo} to generate a video caption emphasizing dynamic information and apply GPT-4o~\cite{Gpt-4o} to generate keyframe caption emphasizing spatial details. For each segmented audio clip, we apply Qwen-Audio-Chat (7B)~\cite{chu2023qwen} to generate an audio caption, and utilize Whisper-Large-V3~\cite{radford2023robust} to get accurate subtitles. 
Note that we found that the performance of the audio captioner lags significantly behind that of visual models, leading to more hallucination issues, such as generating repetitive sentences or incorrect ASR. To address this, we cleaned up these generations, retaining only general descriptions for each audio event (\eg, "this is a man speaking") while removing the specific speech content. Accurate ASR outputs generated by the advanced speech recognition model~\cite{radford2023robust} were used as replacements.
After obtaining modality-specific captions, we instruct Gemini-1.5-Pro~\cite{Gemini} to integrate and correlate them explicitly. The detailed prompts are shown in Fig.~\ref{fig:video_cap}.
In Sec.{\color{blue}3.5}, we quantitatively identify the characteristics of our omni-modal event captions, including audio-visual correlations and fine-grained temporal dynamics using Gemini-1.5-Pro~\cite{Gemini}. Here, we provide the detailed prompt as shown in Fig~\ref{fig:avc}.

\section{Task, Model and Training Data Details}
\subsection{Detailed Task Definition}
We extend three fine-grained video tasks to the novel omni-modality domain towards omni-perception of long videos.
These tasks emphasize cross-modal reasoning and fine-grained temporal understanding at the same time. Here, we provide detailed definitions for these tasks.

\noindent\textbf{Omni-modal temporal video grounding.} Given a textual query describing a specific omni-modal event, the model is required to identify the start and end timestamps of the corresponding video segment.

\noindent\textbf{Omni-modal dense video captioning.} The task is more intricate, requiring the model to perform both temporal localization and captioning for all omni-modal events occurring in a given untrimmed video.

\noindent\textbf{Omni-modal segment captioning.} Given a temporal boundary, the task demands the model to generate a caption summarizing the content of the corresponding omni-modal event within the untrimmed video.

\subsection{Detailed Model Architecture}
\noindent\textbf{Multimodal encoders.} Given a video, we utilize a frozen CLIP ViT-L/14~\cite{radford2021learning} as the Visual Encoder to extract visual embeddings $F_{V}=\{v_{i}\}^{N_{v}}_{i=1}$, where $N_{v}$ denotes the number of input video frames. 
Since both non-speech audio (\ie, natural sound and music) and speech provide crucial information for multi-modal video understanding, we employ BEATs~\cite{chen2022beats} and Whisper~\cite{radford2023robust} to extract non-speech audio embeddings $F_{A}=\{a_{i}\}_{i=1}^{N_{a}}$ and speech embeddings $F_{S}=\{s_{i}\}_{i=1}^{N_{s}}$, where $N_{a}$ and $N_{s}$ represent the number of audio and speech embeddings, respectively.
Therefore, the resulting auditory features of these two encoders are complementary and suitable for general audio input.

\noindent\textbf{Multimodal adapters.} We apply linear layers to project the obtained embeddings from different modalities to get visual tokens $\hat{F_{V}}=\{\hat{v_{i}}\}^{N_{v}}_{i=1}$, audio tokens $\hat{F_{A}}=\{\hat{a_{i}}\}_{i=1}^{N_{a}}$, and speech tokens $\hat{F_{S}}=\{\hat{s_{i}}\}_{i=1}^{N_{s}}$ that are aligned with LLM's token space. Subsequently, the obtained token sequences are simply concatenated as:
\begin{equation}
    \mathbf{Z} = \mathrm{Concat}(\hat{F_{V}}, \hat{F_{A}}, \hat{F_{s}}),
\end{equation}
where $\mathbf{Z}\in \mathbb{R}^{N\times d}$, $N=N_{v}+N_{a}+N_{s}$, and $d$ is the hidden dimension of LLM. Note that our model also supports single-modal and dual-modal inputs, allowing for flexible handling of video data with missing modalities.

\noindent\textbf{Large language model.} We use Vicuna-7B-v1.5~\cite{chiang2023vicuna} as our LLM to process concatenated multi-modal tokens $\mathbf{Z}$ and user queries for response generation.

\subsection{Training Data Details}
\begin{figure*}[h]
  \centering
  \setlength{\abovecaptionskip}{1.0mm}
   \includegraphics[width=1.0\linewidth]{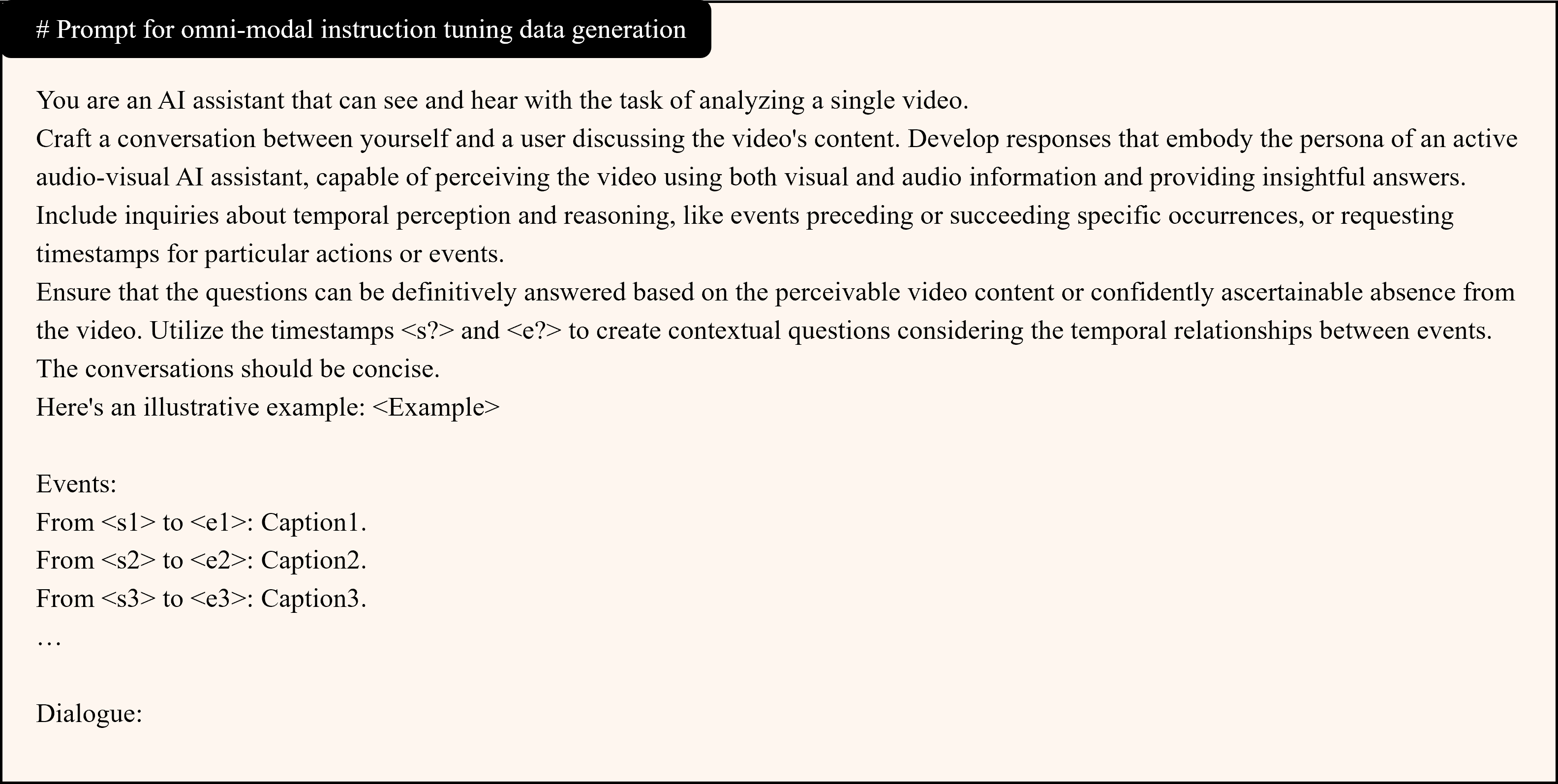}
   \caption{The prompt used to generate omni-modal instruction tuning data.}
   \label{fig:instruction}
\end{figure*}

For boundary perception, we adopted the same template-based data generation strategy as~\cite{huang2024vtimellm} with the same templates, where 20\% of the data is randomly sampled to generate single-turn dialogues (omni-modal dense video captioning), and 80\% is used to generate multi-turn dialogues, \ie, each event is randomly assigned to one of the two tasks (omni-modal temporal video grounding and segment captioning).
For instruction tuning, the prompt used to generate omni-modality dialogues is shown in Fig.~\ref{fig:instruction}.

\section{Experimental Details}
\subsection{More Implementation Details}
We train our model for 2 epochs with a batch size of 128 throughout the two training stages. The AdamW~\cite{loshchilov2017decoupled} optimizer is applied with a cosine learning rate decay and a warm-up period. The learning rate is $1\times 10^{-4}$. The rank in LoRA is 64 with $alpha=128$. Following~\cite{huang2024vtimellm}, we merge the LoRA module trained in the boundary perception stage with the LLM parameters, and then additionally incorporate a new LoRA module for instruction tuning. This ensures the temporal understanding capabilities acquired during the boundary perception stage are effectively preserved within the model.
We complete the training of our 7B model within 30 hours with 1 RTX-A100 (40G) GPU.

\subsection{Evaluation Details}
\noindent\textbf{Evaluation of our LongVALE-LLM.}
For LongVALE-LLM that only undergoes boundary perception tuning without instruction tuning, we directly use the templates as queries. Specifically, for the omni-modal dense captioning task, we employ ``\textit{Could you please detail the events that took place during different time segments in the video?}" as the query. For the omni-modal temporal grounding task, we employ ``\textit{During which frames does $<event>$ occur in the video?}" as the query. For the omni-modal segment captioning task, we employ ``\textit{Could you tell me what happened from $<start>$ to $<end>$ in the video?}" as the query.
LongVALE-LLM that undergoes instruction tuning demonstrates strong instruction-following ability. For omni-modal dense captioning, we utilize the following query: ``\textit{Could you please detail the events that took place during different time segments in the video? List the events in the format: From xx to xx, event1. From xx to xx, event2...}". 
For the omni-modal temporal grounding task, we employ the query ``\textit{During which frames does $<event>$ occur in the video? Give the timestamps in the format: From xx to xx.}" 
or the omni-modal segment captioning task, we employ the query ``\textit{Can you describe what occurred from $<start>$ to 
$<end>$ in the video? Please give the event description directly.}".
We also adopt other similar queries such as ``\textit{Provide details about the events from $<start>$ to $<end>$ in the video...}", the results remain consistently close. 

\noindent\textbf{Evaluation of other video LLMs.}
For other Video LLMs including VideoLLaMA, PandaGPT, NExT-GPT, VideoChat, Video-ChatGPT, TimeChat, and VTimeLLM, we tried our best to assess their optimal performance, recognizing that some were not specifically trained for these tasks. 
For models that have been trained on tasks such as dense video captioning or grounding, we employ the queries provided in their original studies. For instance, for TimeChat, we use the original query for dense captioning: ``\textit{Localize a series of activity events in the video, output the start and end timestamp for each event, and describe each event with sentences. List the events in the format: From x1 second to y1 second: event1.}'' 
Similarly, for temporal grounding, we use the query: ``\textit{Detect and report the start and end timestamps of the video segment that semantically matches the \{sentence\}. Give the timestamps in the format: From xx to xx.}'' For segment captioning, we identified the most effective prompt to be the one described below. 

For models such as VideoLLaMA, PandaGPT, and Video-ChatGPT without training for these tasks, we found that the most effective approach involved using queries that include the video duration. For dense captioning, the query, ``\textit{This video has a duration of D seconds. From which second to which second in the video, what event happens? List the events in the format: From x1 second to y1 second: event1...}'' yielded the best results. 
For grounding, we found that the query, ``\textit{This video lasts for D seconds. During this time, at what specific time does the event \{sentence\} occur? Please provide the start and end timestamps in the format: From x seconds to y seconds, the event happens.}'' produced optimal performance. Moreover, we used GPT-4o mini to efficiently extract timestamps from the generated responses.
Additionally, for segment captioning, we observed that using ``\textit{This video has a total duration of D seconds. Please describe in detail what happens between $<start>$ and $<end>$ in the video. Be specific about the activities of individuals, the environment, and any interactions between people or objects.}'' provided the clearest and most detailed outputs. 
After obtaining the output, we tried to apply multiple regular expressions to format the output. For those outputs cannot be processed, we exclude the corresponding data from metric calculations.

\section{More Qualitative Results}
\begin{figure}[h]
  \centering
  \setlength{\abovecaptionskip}{1.0mm}
   \includegraphics[width=1.0\linewidth]{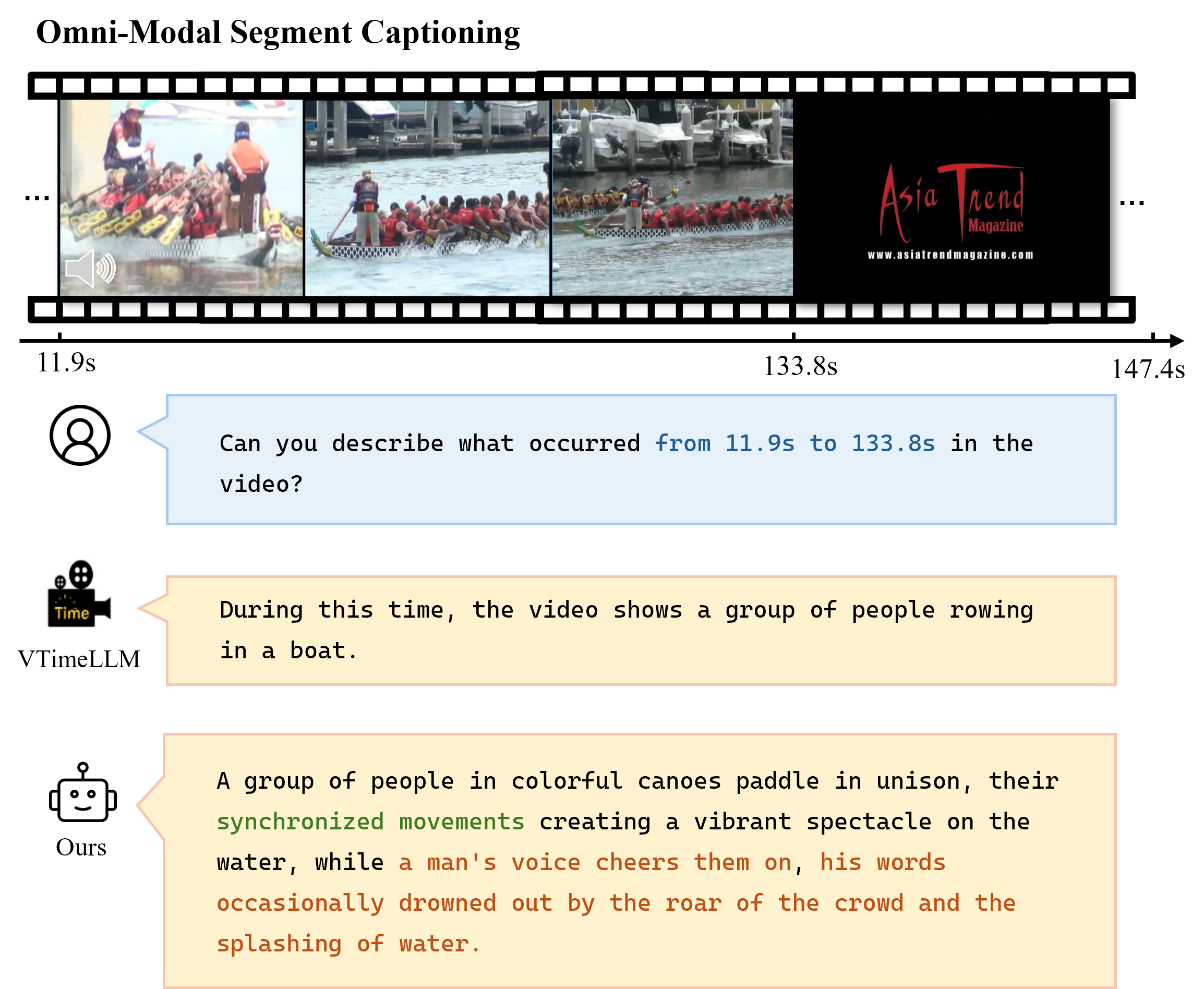}
   \caption{Additional qualitative results on omni-modal segment captioning task. The sample is from LongVALE test set.}
   \label{fig:visual1}
\end{figure}

\begin{figure}[h]
  \centering
  \setlength{\abovecaptionskip}{1.0mm}
   \includegraphics[width=1.0\linewidth]{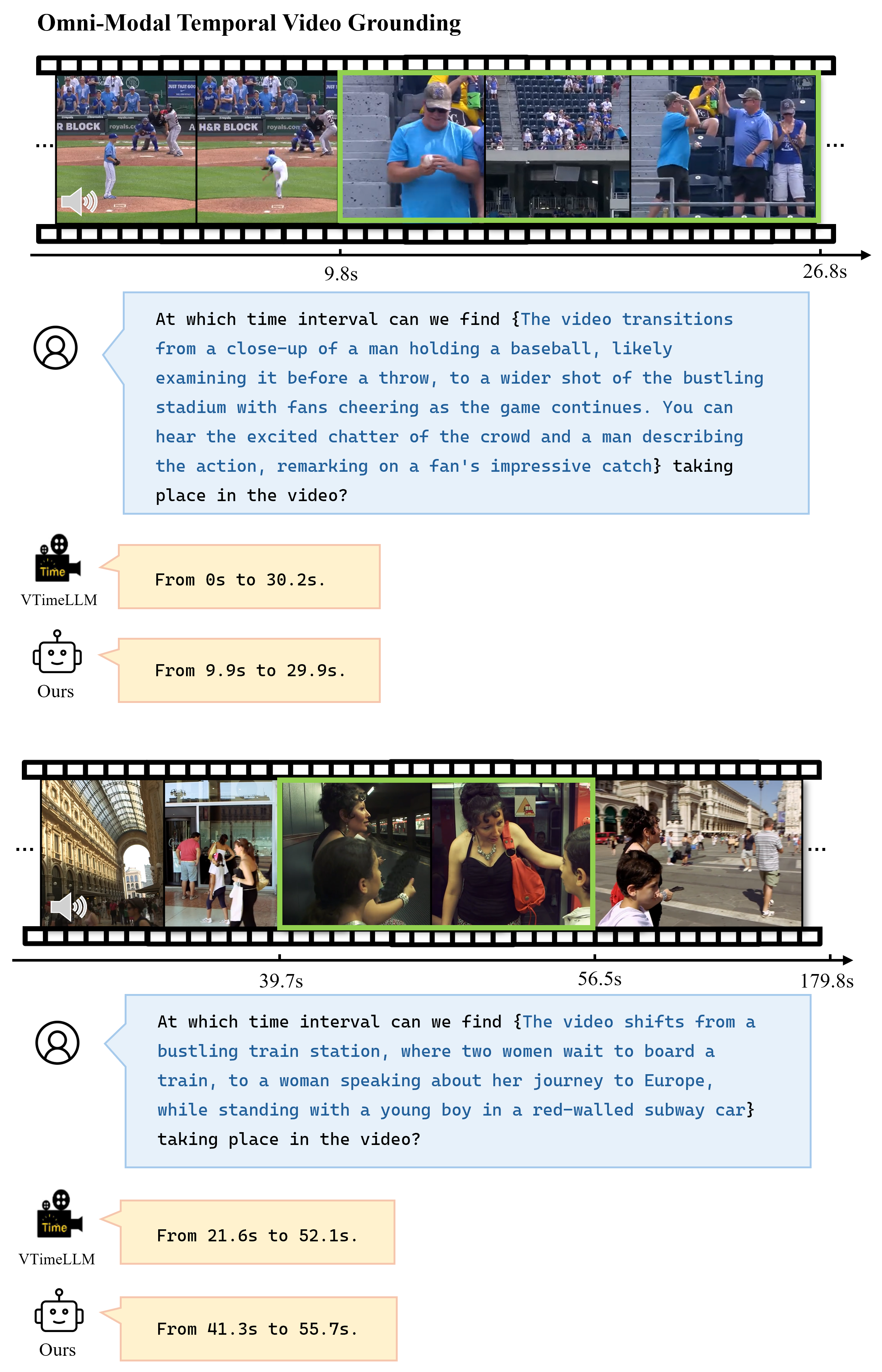}
   \caption{Qualitative results on omni-modal temporal video grounding task. The sample is from LongVALE test set. The ground-truth boundaries are displayed in green. }
   \label{fig:visual2}
\end{figure}

\begin{figure}[h]
  \centering
  \setlength{\abovecaptionskip}{1.0mm}
   \includegraphics[width=1.0\linewidth]{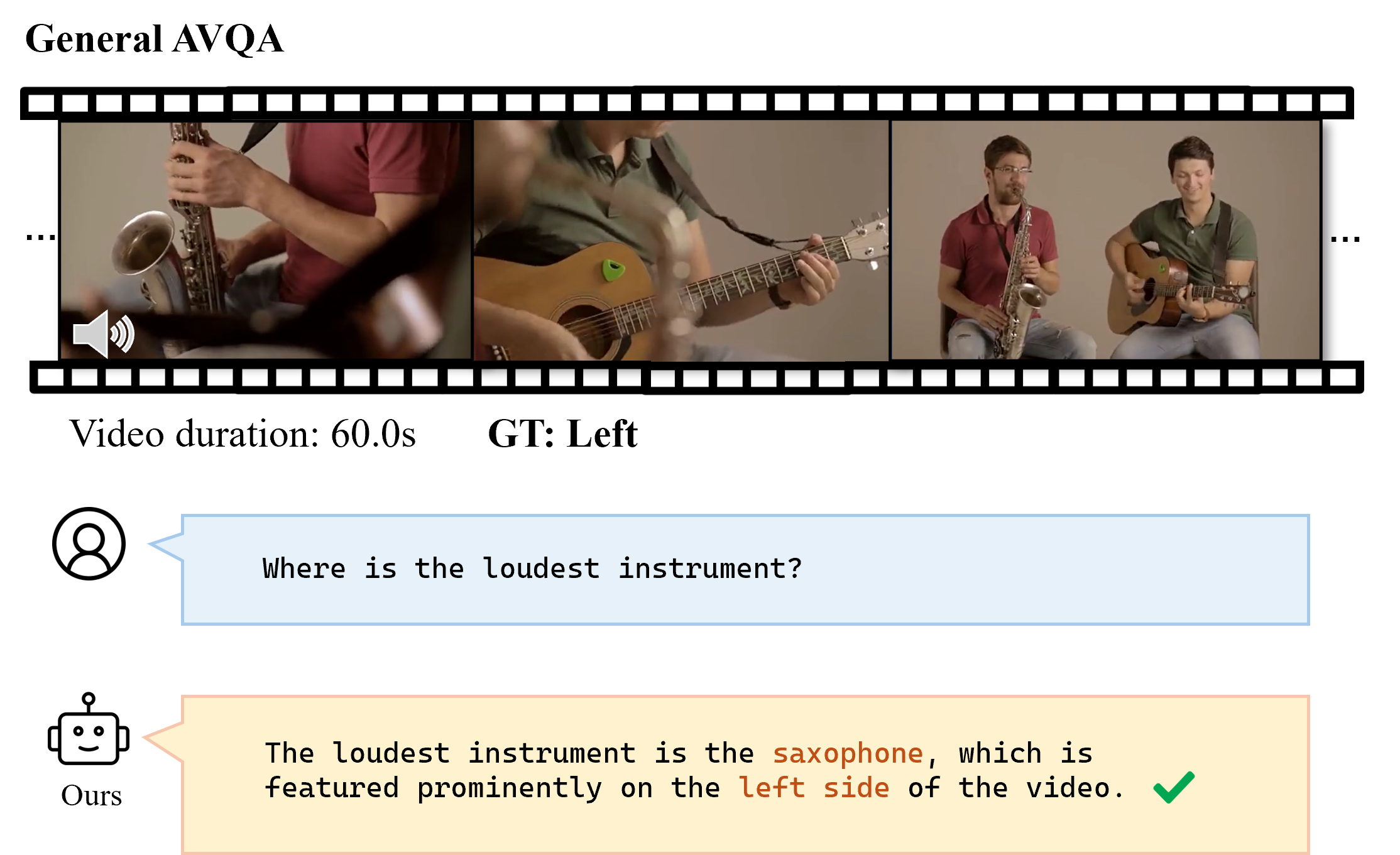}
   \caption{Additional qualitative results on general audio-visual question answering (AVQA) task. The sample is from Music-AVQA test set.}
   \label{fig:visual3}
\end{figure}

\begin{figure}[h]
  \centering
  \setlength{\abovecaptionskip}{1.0mm}
   \includegraphics[width=1.0\linewidth]{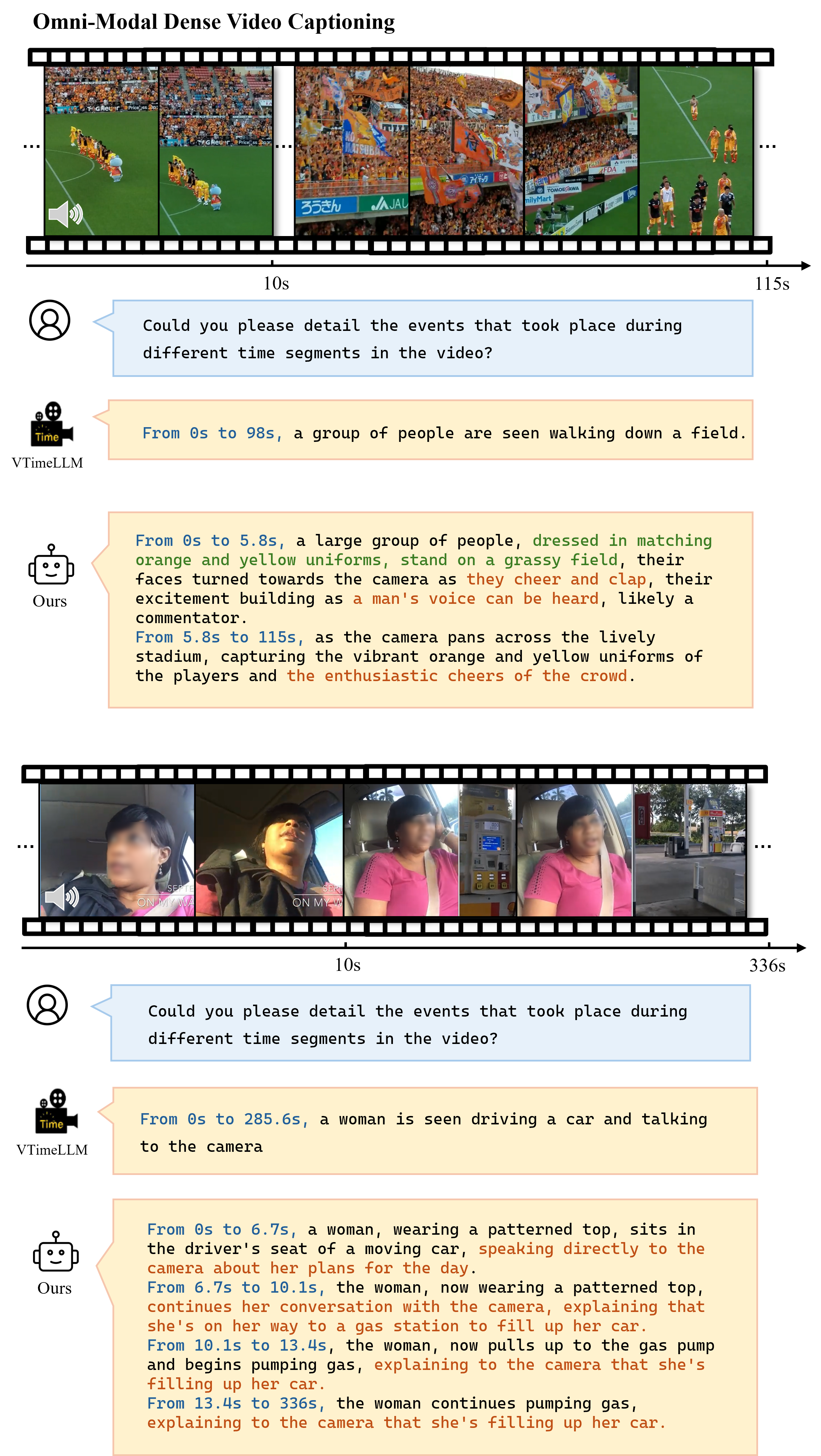}
   \caption{Qualitative results on omni-modal dense video captioning task. The sample is from LongVALE test set.}
   \label{fig:visual4}
\end{figure}

As shown in Fig.~\ref{fig:visual1}-\ref{fig:visual4}, we present more qualitative results encompassing all evaluated tasks. 

\noindent\textbf{Omni-modal segment captioning.} In Fig.~\ref{fig:visual1}, VTimeLLM provides only brief descriptions of visual events within the specified moment, whereas our model offers richer information on both dynamic and auditory events, delivering a more comprehensive and vivid account.

\noindent\textbf{Omni-modal temporal video grounding.} In Fig.~\ref{fig:visual2}, given an omni-modal event caption, our model can more accurately pinpoint the time interval when the event occurs, which fully demonstrates its fine-grained temporal understanding capability in an omni-modality domain. 

\noindent\textbf{Omni-modal dense video captioning.} In Fig.~\ref{fig:visual4}, given a video, our model can identify more omni-modal events and provide finer-grained descriptions, including key information from both visual and audio modalities, enabling a full understanding of the video’s storyline.

\noindent\textbf{General audio-visual question answering (AVQA).} Our model not only excels in fine-grained omni-modal understanding but also demonstrates the ability to accurately answer more general audio-visual questions through cross-modal reasoning. For instance, in Fig.~\ref{fig:visual3}, it can precisely determine the location of the loudest instrument by integrating visual and auditory cues.

Overall, these examples vividly illustrate that relying solely on visual information to understand videos is far from sufficient. Integrating information from multiple modalities is both crucial and essential for comprehensive video understanding. Furthermore, thanks to our LongVALE dataset, our model is the first to combine cross-modal reasoning with fine-grained temporal understanding, setting it apart from traditional vision-only models.

\section{Broader Impact}
\noindent\textbf{Risk mitigation.}
During the data generation, we used Gemini's safety mechanism to efficiently block harmful responses (\ie, harassment, hate, dangerous content, \etc) and filter out corresponding videos. We also removed all individual names with the NLTK framework to protect privacy.

\noindent\textbf{Data Licenses.} We sourced our data from the open-sourced database, ACAV-100M~\cite{lee2021acav100m} under MIT License\footnote{https://opensource.org/license/mit}.
Besides, the annotations of our LongVALE will be provided to the public under CC BY-NC-SA 4.0 license\footnote{https://creativecommons.org/licenses/by-nc-sa/4.0/}. 
We hope our dataset can serve as a pivotal benchmark for promoting comprehensive multi-modal video understanding.

\end{document}